\documentclass[10pt,journal,compsoc,final]{IEEEtran}
\ifCLASSINFOpdf
\else
\fi
\usepackage{graphicx,subfigure,subfig,color}
\usepackage{bm,amsmath,amssymb,amsthm,dsfont}
\usepackage{algorithm,algorithmic}
\usepackage{url}
\usepackage{balance}
\usepackage{threeparttable}

\newtheorem{theorem}{Theorem}

\begin{document}
\title{Group-sparse SVD Models and Their Applications in Biological Data}
\author{Wenwen Min, Juan Liu and Shihua Zhang
\IEEEcompsocitemizethanks{
\IEEEcompsocthanksitem Wenwen Min is with School of Mathematics and Computer Science, Jiangxi Science and Technology Normal University, Nanchang 330038, China. E-mail: minwenwen07@163.com
\IEEEcompsocthanksitem Juan Liu is with School of Computer Science, Wuhan University, Wuhan 430072, China. E-mail: liujuan@whu.edu.cn.
\IEEEcompsocthanksitem Shihua Zhang is with NCMIS, CEMS, RCSDS, Academy of Mathematics and Systems Science, Chinese Academy of Sciences, Beijing 100190; School of Mathematical Sciences, University of Chinese Academy of Sciences, Beijing 100049; Center for Excellence in Animal Evolution and Genetics, Chinese Academy of Sciences, Kunming 650223, China. E-mail: zsh@amss.ac.cn.}
\thanks{Manuscript received XXX, 2018; revised XXX, 2018.}}
\markboth{ }
{Min \MakeLowercase{\textit{et al.}}: Network-regularized Sparse Logistic Regression Models for Clinical Risk Prediction and Biomarker Discovery}

\IEEEtitleabstractindextext{
\begin{abstract}
  Sparse Singular Value Decomposition (SVD) models have been proposed for biclustering high dimensional gene expression data to identify block patterns with similar expressions. However, these models do not take into account prior group effects upon variable selection. To this end, we first propose group-sparse SVD models with group Lasso ($GL_1$-SVD) and group $L_0$-norm penalty ($GL_0$-SVD) for non-overlapping group structure of variables. However, such group-sparse SVD models limit their applicability in some problems with overlapping structure. Thus, we also propose two group-sparse SVD models with overlapping group Lasso ($OGL_1$-SVD) and overlapping group $L_0$-norm penalty ($OGL_0$-SVD). We first adopt an alternating iterative strategy to solve $GL_1$-SVD based on a block coordinate descent method, and $GL_0$-SVD based on a projection method. The key of solving $OGL_1$-SVD is a proximal operator with overlapping group Lasso penalty. We employ an alternating direction method of multipliers (ADMM) to solve the proximal operator. Similarly, we develop an approximate method to solve $OGL_0$-SVD. Applications of these methods and comparison with competing ones using simulated data demonstrate their effectiveness. Extensive applications of them onto several real gene expression data with gene prior group knowledge identify some biologically interpretable gene modules.

\end{abstract}
}
\maketitle
\IEEEpeerreviewmaketitle

\section{Introduction}
\IEEEPARstart{S}{ingular} Value Decomposition (SVD) is one of the classical matrix decomposition models \cite{singh2008unified}. It is a useful tool for data analysis and low-dimensional data representation in many different fields such as signal processing, matrix approximation and bioinformatics \cite{alter2000singular,aharon2006img,zhou2015svd}. However, the non-sparse singular vectors with all variables are difficult to be explained intuitively. In the recent years, sparse models have been widely applied in computational biology to improve biological interpretation \cite{sohn2009gradient,fang2015cclasso,greenlaw2017bayesian}. In addition, many researchers applied diverse sparse penalties onto singular vectors in SVD and developed multiple sparse SVD models to improve their interpretation and capture inherent structures and patterns from the input data \cite{shi2013scmf,jing2015sparse}. For example, sparse SVD provides a new way for exploring bicluster patterns of gene expression data. Suppose $\bm{X} \in \mathbb{R}^{p\times n}$ denotes a gene expression matrix with $p$ genes and $n$ samples. Biologically, a subset of patients and genes can be clustered together as a coherent bicluster or block pattern with similar expressions. Previous studies have reported that such a bicluster among gene expression data can be identified by low-rank sparse SVD models \cite{lee2010biclustering,sill2011robust,yang2014sparse}. However, these sparse models ignore prior information of gene variables, and usually assume that each gene is selected in a bicluster with equal probability. Actually, one gene may belong to multiple pathways in biology \cite{kanehisa2000kegg}. As far as we know, there is not yet a model for biclustering gene expression data by integrating gene pathway information. Group sparse penalties \cite{yuan2006model,jacob2009group} should be used to induce the structured sparsity of variables for variable selection. Several studies have explored the (overlapping) group Lasso in regression tasks \cite{obozinski2011group,yuan2011efficient}. However, little work focus on developing structured sparse SVD for biclustering high-dimensional data (e.g., biclustering gene expression data via integrating prior gene group knowledge).

In this paper, motivated by the development of sparse coding and structured sparse penalties, we propose several group-sparse SVD models for pattern discovery in biological data. We first introduce the group-sparse SVD model with group Lasso ($L_1$) penalty ($GL_1$-SVD) to integrate non-overlapping structure of variables. Compared to $L_1$-norm, $L_0$-norm is a more natural sparsity-inducing penalty. Thus, we also propose an effective group-sparse SVD via replacing $L_1$-norm with $L_0$-norm, called $GL_0$-SVD, which uses a mix-norm by combining the group Lasso and $L_0$-norm penalty. However, the non-overlapping group structure limits their applicabilities in diverse fields. We consider a more general situation, where we assume that either groups of variables are potentially overlapping (e.g., a gene may belong to multiple pathways (groups)). We also propose two group-sparse SVD models with overlapping group Lasso ($OGL_1$-SVD) and overlapping group $L_0$-norm penalty ($OGL_0$-SVD).

To solve these models, we design an alternating iterative  algorithm to solve $GL_1$-SVD based on a block coordinate descent method and $GL_0$-SVD based on a projection method. Furthermore, we develop a more general approach based on Alternating Direction Method of Multipliers (ADMM) to solve $OGL_1$-SVD. In addition, we extend $OGL_1$-SVD to $OGL_0$-SVD, which is a regularized SVD with overlapping grouped $L_0$-norm penalty. The key of solving $OGL_1$-SVD is also a proximal operator with overlapping group $L_0$-norm penalty. We propose a greedy method to solve it and obtain its approximate solution. Finally, applications of these methods and comparison with the state-of-the-art ones using a set of simulated data demonstrate their effectiveness and computational efficiency. Extensive applications of them onto the high-dimensional gene expression data show that our methods could identify more biologically relevant gene modules, and improve their biological interpretations.

\textbf{Related Work} We briefly review the regularized low rank-$r$ SVD model as follows:
\begin{equation}
\begin{aligned}\label{equ:SVD}
& \underset{\bm{U}, \bm{D}, \bm{V}}{\text{minimize}} && \|\bm{X} - \bm{UDV}^T\|_F^2\\
& \text{subject to}                  && \|\bm{U}_i\|^2 \leq 1, \Omega_1(\bm{U}_i) \leq c_1^{i}, \forall i\\
&                                    && \|\bm{V}_i\|^2 \leq 1, \Omega_2(\bm{V}_i) \leq c_2^{i}, \forall i
\end{aligned}
\end{equation}
where $\bm{X} \in \mathbb{R}^{p\times n}$ with $p$ features and $n$ samples, $\bm{U}\in \mathbb{R}^{p\times r}$, $\bm{V}\in \mathbb{R}^{r\times n}$ and $\bm{D}$ is diagonal matrix. $\bm{U}_i$ ($\bm{V}_i$) corresponds to the $i$-th column of $\bm{U}$ ($\bm{V}$), which is a column orthogonal matrix. To solve the above optimization problem, we introduce a general regularized rank-one SVD model:
\begin{equation}
\begin{aligned}\label{equ:01}
& \underset{\bm{u},\bm{v},d}{\text{minimize}} && \|\bm{X} - d\bm{uv}^T\|_F^2\\
& \text{subject to}                  && \|\bm{u}\|^2 \leq 1, \Omega_1(\bm{u}) \leq c_1,\\
&                                    && \|\bm{v}\|^2 \leq 1, \Omega_2(\bm{v}) \leq c_2,
\end{aligned}
\end{equation}
where $d$ is a positive singular value, $\bm{u}$ is a $p$-dimensional column vector, and $\bm{v}$ is a $n$-dimensional column vector. $\Omega_1(\bm{u})$ and $\Omega_2(\bm{v})$ are two penalty functions, $c_1$ and $c_2$ are two hyperparameters. In a Bayesian view, different prior distribution functions of $\bm{u}$ and $\bm{v}$ correspond to different regularized functions. For example, $L_1$-norm is a very popular sparsity-inducing norm \cite{tibshirani1996regression} and has been used to obtain sparse solutions in a large number of statistical models including the regression model \cite{tibshirani1996regression,zou2005regularization}, SVD \cite{witten2009penalized}, PCA \cite{zou2006sparse}, LDA \cite{clemmensen2011sparse}, K-means \cite{witten2010framework}, etc.

Recently, some sparse SVD models have been proposed for coherent sub-matrix detection \cite{witten2009penalized,lee2010biclustering,sill2011robust}. For example, Witten \emph{et~al.} \cite{witten2009penalized} developed a penalized matrix decomposition (PMD) method, which regularizes the singular vectors with Lasso and fussed Lasso to induce sparsity. Lee \emph{et~al.} \cite{lee2010biclustering} proposed a rank-one sparse SVD model with adaptive Lasso ($L_1$) ($L_1$-SVD) of the singular vectors for biclustering of gene expression data. Some generalized sparsity penalty functions (e.g., group Lasso \cite{yuan2006model} and sparse group lasso \cite{Noah2013A}) have been widely used in many regression models for feature selection by integrating group information of variables. However, it is a challenging issue to use these generalized penalty functions such as group Lasso and overlapping group Lasso \cite{jacob2009group,tibshirani2011regression} in the SVD framework with effective algorithms. To this end, we develop several group-sparse SVD models with different group-sparse penalties including $\Omega_{GL_1}(\bm{u})$, $\Omega_{GL_0}(\bm{u})$, $\Omega_{OGL_1}(\bm{u})$ and $\Omega_{OGL_0}(\bm{u})$ to integrate diverse group structures of variables for pattern discovery in biological data (see TABLE 1).
\begin{figure}[htbp]
  \centering
  \includegraphics[width=1.01\linewidth]{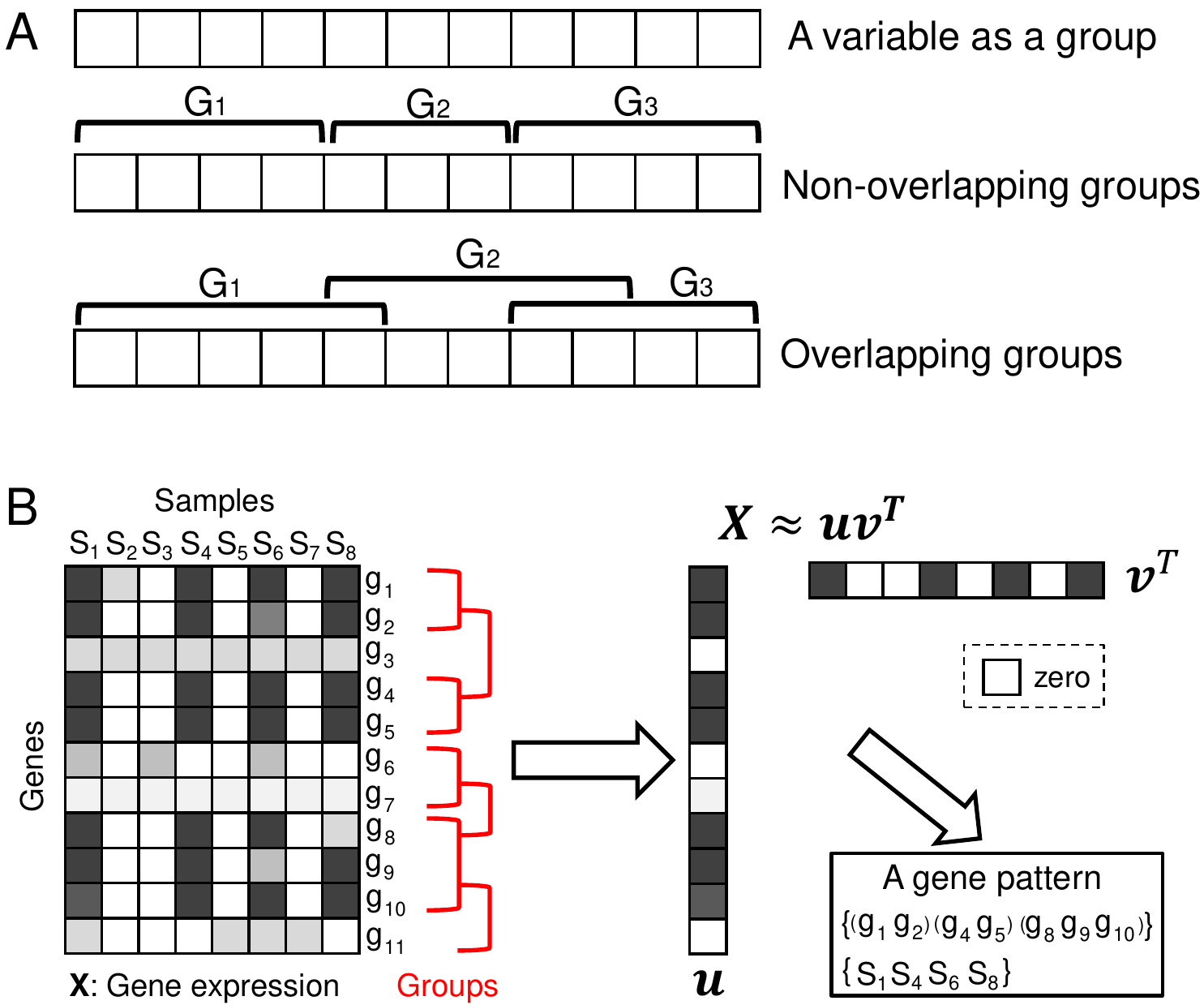}
  \caption{(A) There are three structured ways corresponding to three penalty functions for variable selection including Lasso, group Lasso and overlapping group Lasso. In the third way, group $G_1$ and $G_2$ as well as group $G_2$ and $G_3$ are overlapping, respectively. (B) A simple example to explain how SVD($OGL_0$,$L_0$) (see TABLE 1) identifies a gene co-expression sub-matrix by integrating gene expression data and prior group information. Genes are divided into seven overlapping groups (denoted by brackets). SVD($OGL_0$,$L_0$) is used to identify a pair sparse singular vectors $\bm{u}$ and $\bm{v}$. Based on the non-zero elements of $\bm{u}$ and $\bm{v}$, a gene bicluster or module can be identified with a gene set $\{(g_1,g_2),(g_4,g_5),(g_8,g_9,g_{10}))\}$ and a sample set $\{s_1,s_4,s_6,s_8\}$, where ``( )" indicates some genes are in a given group.}\label{fig01}
\end{figure}

\section{Group-sparse SVD Models}
In this section, we propose four group sparse SVD models with respect to different structured penalties (TABLE 1). For a given data (e.g., gene expression data), we can make proper adjustments to get one-sided group-sparse SVD models via using (overlapping) group-sparse penalties for the right (or left) singular vector. For example, SVD($OGL_0$, $L_0$) is a group-sparse SVD model, which uses the overlapping group $L_0$-penalty for $\bm{u}$ and $L_0$-penalty for $\bm{v}$ respectively.

\begin{table}[ht]
\centering
\begin{threeparttable}\label{table:1}
\caption{The group-sparse SVD models}
\begin{tabular}{l|l}
   \hline
   \textbf{Model}  & \textbf{Penalty function}\\
   \hline
   $GL_1$-SVD  & Group-Lasso ($GL_1$)\\
   $GL_0$-SVD  & Group-$L_0$ ($GL_0$)\\
   $OGL_1$-SVD & Overlapping-Group-Lasso ($OGL_1$)\\
   $OGL_0$-SVD & Overlapping-Group-$L_0$ ($OGL_0$)\\
   \hline
\end{tabular}
\end{threeparttable}
\end{table}

Below we will introduce these models and their algorithms in detail.
\subsection{$\bm{GL_1}$-SVD}
Suppose the left singular vector $\bm{u}$ and right singular vector $\bm{v}$ can be respectively divided into $L$ and $M$ non-overlapping groups: $\bm{u}^{(l)}\in \mathbb{R}^{p_l\times 1}, l = 1,...,L$ and $\bm{v}^{(m)}\in \mathbb{R}^{p_m\times 1}, m = 1,...,M$. Here, we consider the (adaptive) group Lasso ($GL_1$) penalty \cite{Wang2008A} for $\bm{u}$ and $\bm{v}$ as follows:
\begin{equation}\label{equ:GL1}
  \Omega_{GL_1}(\bm{u}) = \sum_{l=1}^L w_l \|\bm{u}^{(l)}\|_2~\mbox{and}~\Omega_{GL_1}(\bm{v}) = \sum_{m=1}^M \tau_m \|\bm{v}^{(m)}\|_2,
\end{equation}
where both $w_l$ and $\tau_m$ are adaptive weight parameters. Suppose $w_l = \sqrt{p_l}$ and $\tau_m = \sqrt{q_m}$ for group sizes, the penalty reduces to a traditional group Lasso.

Based on the definition of $GL_1$ penalty, we propose the first group-sparse SVD with group Lasso penalty ($GL_1$-SVD), also namely SVD($GL_1$, $GL_1$):
\begin{equation}
\begin{aligned}\label{equ:02}
& \underset{\bm{u},\bm{v},d}{\text{minimize}} && \|\bm{X} - d\bm{uv}^T\|_F^2\\
& \text{subject to}                         && \|\bm{u}\|^2 \leq 1, \Omega_{GL_1}(\bm{u}) \leq c_1,\\
&                                           && \|\bm{v}\|^2 \leq 1, \Omega_{GL_1}(\bm{v}) \leq c_2.
\end{aligned}
\end{equation}
Since $||\bm{X}-d\bm{u}\bm{v}^T||_F^2 = \|\bm{X}\|_F^2 + d^2 - 2d\bm{u}^T\bm{X}\bm{v}$. Minimizing $||\bm{X}-d\bm{u}\bm{v}^T||_F^2$ is equivalent to minimizing $-\bm{u}^T\bm{X}\bm{v}$, and once the $\bm{u}$ and $\bm{v}$ are determined, the $d$ value is determined by $\bm{u}^T\bm{X}\bm{v}$. We obtain the Lagrangian form of $GL_1$-SVD model as follows:
\begin{equation}\label{equ:03}
\begin{split}
\mathcal{L}(\bm{u},\bm{v})  = & -\bm{u}^T\bm{X}\bm{v} + \lambda_1\Omega_{GL_1}(\bm{u}) + \lambda_2\Omega_{GL_1}(\bm{v})  \\
                              & + \eta_1\|\bm{u}\|^2  + \eta_2\|\bm{v}\|^2,
\end{split}
\end{equation}
where $\lambda_1\geq 0$, $\lambda_2 \geq 0$, $\eta_1\geq 0$ and $\eta_2\geq 0$ are Lagrange multipliers. To solve the problem (\ref{equ:03}), we apply an alternating iterative algorithm to optimize $\bm{u}$ for a fixed $\bm{v}$ and \emph{vice versa}.

\subsubsection{Learning $\bm{u}$}
Fix $\bm{v}$ and let $\bm{z}=\bm{X}\bm{v}$, minimizing Eq. (\ref{equ:03}) is equivalent to minimizing the following criterion:
\begin{equation}\label{equ:04}
  \mathcal{L}(\bm{u},\lambda,\eta) = -\bm{u}^T\bm{z} + \lambda \sum_{l=1}^L w_l \|\bm{u}^{(l)}\|_2+ \eta \sum_{l=1}^L {\bm{u}^{(l)}}^T\bm{u}^{(l)}  ,
\end{equation}
where $\bm{u} = [\bm{u}^{(1)};\bm{u}^{(2)}; \dots ;\bm{u}^{(L)}]$ and $\lambda = \lambda_1$, $\eta = \eta_1$ for simplicity. It is obvious that $\mathcal{L}(\bm{u},\lambda,\eta)$ is convex with respect $\bm{u}$, and we develop a block coordinate descent algorithm \cite{Tseng2001Convergence,friedman2007pathwise,friedman2010regularization,simon2013blockwise} to minimize Eq. (\ref{equ:04}), i.e. one group of $\bm{u}$ is updated at a time. For a single group $\bm{u}^{(l)}$ with fixed $\bm{u}^{(j)}$ for all $1\leq j \leq L$ and $j \neq l$, the subgradient equations (see \cite{bazaraa2013nonlinear}) of Eq. (\ref{equ:04}) with respect to $\bm{u}^{(l)}$ is written as:
\begin{equation}\label{equ:05}
  \nabla_{\bm{u}^{(l)}}\mathcal{L} = -\bm{z}^{(l)} + \lambda w_l \bm{s}^{(l)} + 2\eta \bm{u}^{(l)} = \bm{0},
\end{equation}
where $\bm{s}^{(l)}$ is the subgradient vector of $\bm{u}^{(l)}$ and it meets
\begin{eqnarray}\label{equ:06}
  \bm{s}^{(l)}=
  \begin{cases}
    \frac{\bm{u}^{(l)}}{\|\bm{u}^{(l)}\|_2}, &\mbox{if}~\bm{u}^{(l)} \neq \bm{0},\cr
    \in \{\bm{s}^{(l)}:~ \|\bm{s}^{(l)}\|_2\leq 1 \}, &\mbox{otherwise}.
  \end{cases}
\end{eqnarray}
Based on Eq. (\ref{equ:05}), we have $2\eta \bm{u}^{(l)}=\bm{z}^{(l)} - \lambda w_l \bm{s}^{(l)}$.

If $\|\bm{z}^{(l)}\|_2 > \lambda w_l$, then we have $\bm{u}^{(l)} \neq 0$. Since $\eta>0$,~$\lambda>0$,~$w_l>0 $ and $2\eta \bm{u}^{(l)}=\bm{z}^{(l)} - \lambda w_l \frac{\bm{u}^{(l)}}{\|\bm{u}^{(l)}\|_2}$. Thus, we have $\frac{\bm{u}^{(l)}}{\|\bm{u}^{(l)}\|_2} = \frac{\bm{z}^{(l)}}{\|\bm{z}^{(l)}\|_2}$ and $\bm{u}^{(l)} = \frac{1}{2\eta}\bm{z}^{(l)} ( 1- \frac{\lambda w_l}{\|\bm{z}^{(l)}\|_2} )$.

If~$\|\bm{z}^{(l)}\|_2 \leq \lambda$, then $\bm{u}^{(l)} =\bm{0}$. In short, we obtain the following update rule for $\bm{u}^{(l)}$ ($l = 1,\cdots, L$),
\begin{eqnarray}\label{equ:07}
  \bm{u}^{(l)}=
  \begin{cases}
    \frac{1}{2\eta}(1- \frac{\lambda w_l}{\|\bm{z}^{(l)}\|_2})\bm{z}^{(l)}, &\mbox{if}~\|\bm{z}^{(l)}\|_2 > \lambda w_l,\cr
    \bm{0}, &\mbox{otherwise}.
  \end{cases}
\end{eqnarray}
Since Eq. (\ref{equ:04}) is strictly convex and \emph{separable}, the block coordinate descent algorithm must converge to its optimal solution \cite{Tseng2001Convergence}. Finally, we can choose an $\eta$ to guarantee $\bm{u} = \frac{\bm{u}}{\|\bm{u}\|_2}$ (normalizing condition).

\subsubsection{Learning $\bm{v}$}
In the same manner, we fix $\bm{u}$ in Eq. (\ref{equ:03}) and let $\bm{z}=\bm{X}^T\bm{u}$. Similarly, we can also obtain the coordinate update rule for $\bm{v}^{(m)}, m=1,2,\cdots,M$.
\begin{eqnarray}\label{equ:08}
  \bm{v}^{(m)}=
  \begin{cases}
    \frac{1}{2\eta }( 1- \frac{\lambda \tau_m}{\|\bm{z}^{(m)}\|_2})\bm{z}^{(m)}, &\mbox{if}~\|\bm{z}^{(m)}\|_2 > \lambda \tau_m,\cr
    \bm{0}, &\mbox{otherwise}.
  \end{cases}
\end{eqnarray}
Furthermore, to meet the normalizing condition, we chose an $\eta$ to guarantee $\bm{v} = \frac{\bm{v}}{\|\bm{v}\|_2}$. Besides, if here each group only contains one element, then the group Lasso penalty reduces to the Lasso penalty. Accordingly, we  get another update formula:
\begin{eqnarray}\label{equ:09}
  \bm{v}_i=
  \begin{cases}
    \frac{1}{2\eta }(1- \frac{\lambda}{\|\bm{z}_i\|_2})\bm{z}_i, &\mbox{if}~|\bm{z}_i| > \lambda,\cr
    \bm{0}, &\mbox{otherwise}.
  \end{cases}
\end{eqnarray}

\subsubsection{$GL_1$-SVD Algorithm}
Based on Eqs. (\ref{equ:07}) and (\ref{equ:08}), we propose an alternating iterative algorithm (Algorithm 1) to solve the $GL_1$-SVD model and its time complexity is $\mathcal{O}(Tnp + Tp^2 + Tn^2)$, where $T$ is the number of iterations. We can control the iteration by monitoring the change of $d$.

In order to display the penalty function for left and right singular vectors, $GL_1$-SVD can also be written in another form SVD($GL_1$, $GL_1$), denoting that the left singular vector $\bm{u}$ is regularized by $GL_1$ penalty and the right singular vector $\bm{v}$ is regularized by $GL_1$ penalty, respectively. Similarly, we can simply modify Algorithm 1 to solve SVD($GL_1$, $L_1$) model, which applies Lasso as the penalty for $\bm{v}$.
\begin{algorithm}[h]
\caption{$GL_1$-SVD or SVD($GL_1$, $GL_1$)} \label{alg:Framwork2}
\begin{algorithmic}[1]
\REQUIRE Matrix $\bm{X} \in \mathbb{R}^{p\times n}$, $\lambda_u$ and $\lambda_v$; Group information
\ENSURE $\bm{u}$, $\bm{v}$ and $d$
\STATE Initialize $\bm{v}$ with $\|\bm{v}\|=1$
\REPEAT
\STATE Let $\bm{z}=\bm{Xv}$
\FOR {$l = 1$~to~$L$}
\IF{$\|\bm{u}^{(l)}\|_2 \leq \lambda_u w_l$}
\STATE $\bm{u}^{(l)} =\bm{0}$
\ELSE
\STATE $\bm{u}^{(l)} = \bm{z}^{(l)} ( 1- \frac{\lambda_u w_l}{\|\bm{z}_l\|_2} )$
\ENDIF
\ENDFOR
\STATE $\bm{u} = \frac{\bm{u}}{\|\bm{u}\|_2}$
\FOR {$m = 1$~to~$M$}
\STATE Let $\bm{z}=\bm{X}^T\bm{u}$
\IF{$\|\bm{v}^{(m)}\|_2 \leq \lambda_v w_m$}
\STATE $\bm{v}^{(m)} =\bm{0}$
\ELSE
\STATE $\bm{v}^{(m)} = \bm{z}^{(m)} ( 1- \frac{\lambda_v w_m}{\|\bm{z}^{(m)}\|_2} )$
\ENDIF
\ENDFOR
\STATE $\bm{v} = \frac{\bm{v}}{\|\bm{v}\|_2}$
\STATE $d=\bm{z}^T\bm{v}$
\UNTIL $d$ convergence
\RETURN $\bm{u}$, $\bm{v}$ and $d$
\end{algorithmic}
\end{algorithm}

\subsection{$\bm{GL_0}$-SVD}
Unlike $GL_1$ penalty, below we consider a group $L_0$-norm penalty ($GL_0$) of $\bm{u}$ and $\bm{v}$ as follows:
\begin{equation}\label{equ:GL1}
  \Omega_{GL_0}(\bm{u}) = \|\phi(\bm{u})\|_0~\mbox{and}~\Omega_{GL_0}(\bm{v}) = \|\phi(\bm{v})\|_0,
\end{equation}
where $\phi(\bm{u}) = [\|\bm{u}^{(1)}\|,~\|\bm{u}^{(2)}\|,\cdots,\|\bm{u}^{(L)}\|]^T$ and $\phi(\bm{v}) = [\|\bm{v}^{(1)}\|,~\|\bm{u}^{(2)}\|, \cdots, \|\bm{v}^{(M)}\|]^T$.

Based on the above definition of $GL_0$ penalty, we propose the second group-sparse SVD model with $GL_0$ penalty, namely $GL_0$-SVD or SVD($GL_0$, $GL_0$):
\begin{equation}
\begin{aligned}\label{equ:L0gSVD}
& \underset{\bm{u},\bm{v},d}{\text{minimize}} && \|\bm{X} - d\bm{uv}^T\|_F^2\\
& \text{subject to}                         && \|\bm{u}\|_2 \leq 1, \Omega_{GL_0}(\bm{u}) \leq k_u,\\
&                                           && \|\bm{v}\|_2 \leq 1, \Omega_{GL_0}(\bm{v}) \leq k_v.
\end{aligned}
\end{equation}
Here, we employ an alternating iterative strategy to solve problem (\ref{equ:L0gSVD}). Fix $\bm{u}$ (or $\bm{v}$), the problem (\ref{equ:L0gSVD}) reduces to a projection problem with group $L_0$-norm penalty.

\subsubsection{Learning $\bm{u}$}
Since $||\bm{X}-d\bm{u}\bm{v}^T||_F^2 = \|\bm{X}\|_F^2 + d^2 - 2d\bm{u}^T\bm{X}\bm{v}$. Fix $\bm{v}$ and let $\bm{z}_u= \bm{X}\bm{v}$, Eq. (\ref{equ:L0gSVD}) reduces to a group-sparse projection operator with respect to $\bm{u}$:
\begin{equation}\label{equ:gsp}
  \underset{\|\bm{u}\| \leq 1}{\text{minimize}}~-\bm{z}_u^T\bm{u},~~\mbox{s.t.}~\Omega_{GL_0}(\bm{u}) \leq k_u.
\end{equation}
We present Theorem 1 to solve problem (\ref{equ:gsp}).
\begin{theorem}
The optimum solution of Eq. (\ref{equ:gsp}) is $\frac{\mathcal{P}_{GL_0}(\bm{z}_u)}{\|\mathcal{P}_{GL_0}(\bm{z}_u)\|}$, where $\mathcal{P}_{GL_0}(\bm{z}_u)$ is a column-vector and meets
\begin{eqnarray}\label{equ:P-GL0}
  [\mathcal{P}_{GL_0}(\bm{z}_u)]^{(g)}=
  \begin{cases}
    {\bm{z}_u}^{(g)}, &\mbox{if}~g \in \mbox{supp}(\phi(\bm{z}_u),k_u),\cr
    \bm{0}, &\mbox{otherwise},
  \end{cases}
\end{eqnarray}
where $[\mathcal{P}_{GL_0}(\bm{z}_u)]^{(g)}$ is a sub-vector from the $g$-th group, $g = 1,2,\cdots,L$ and $\mbox{supp}(\phi(\bm{z}_u), k_u)$ denotes the set of indexes of the largest $k_u$ elements of $\phi(\bm{z}_u)$.
\end{theorem}
The objective function of (\ref{equ:gsp}) can be simplified as $-{\bm{z}_u}^T\bm{u} = \sum_{l=1}^{L} -{\bm{z}_u^{(l)}}^T\bm{u}^{(l)}$. Theorem 1 shows that solving problem (\ref{equ:gsp}) is equivalent to forcing the elements in $L - k_u$ groups of $\bm{z}_u$ with the smallest group-norm values to be zeros. We can easily prove that Theorem 1 is true. Here we omit the prove process.

\subsubsection{Learning $\bm{v}$}
In the same manner, fix $\bm{u}$ and let $\bm{z}_v= \bm{X}^T\bm{u}$, thus problem (\ref{equ:L0gSVD}) can be written as a similar subproblem with respect to $\bm{v}$:
\begin{equation}\label{equ:gsp2}
  \underset{\|\bm{u}\| \leq 1}{\text{minimize}}~-\bm{z}_v^T\bm{v},~~\mbox{s.t.}~\Omega_{GL_0}(\bm{v}) \leq k_v.
\end{equation}
Similarly, based on Theorem 1, we can obtain the estimator of $\bm{v}$ as $\frac{\mathcal{P}_{GL_0}(\bm{z}_v)}{\|\mathcal{P}_{GL_0}(\bm{z}_v)\|_2}$.

\subsubsection{$GL_0$-SVD Algorithm}
Finally, we propose an alternating iterative method (Algorithm \ref{alg:GL0-SVD}) to solve the optimization problem (\ref{equ:L0gSVD}). The time complexity of Algorithm \ref{alg:GL0-SVD} is $\mathcal{O}(Tnp+Tn^2+Tp^2)$, where $T$ is the number of iterations.
\begin{algorithm}[h]
\caption{$GL_0$-SVD or SVD($GL_0$, $GL_0$)}\label{alg:GL0-SVD}
\begin{algorithmic}[1]
\REQUIRE Matrix $\bm{X}\in \mathbb{R}^{p\times n}$, $k_u$ and $k_v$; Group information
\ENSURE $\bm{u}$, $\bm{v}$ and $d$.
\STATE Initialize $\bm{v}$ with $\|\bm{v}\|=1$
\REPEAT
\STATE Let $\bm{z}_u=\bm{Xv}$
\STATE $\bm{\widehat{u}} = \mathcal{P}_{GL_0}(\bm{z}_u)$ by using Eq. (\ref{equ:P-GL0})
\STATE $\bm{u} = \frac{\bm{\widehat{u}}}{\|\bm{\widehat{u}}\|_2}$
\STATE Let $\bm{z}_v=\bm{X}^T\bm{u}$
\STATE $\bm{\widehat{v}} = \mathcal{P}_{GL_0}(\bm{z}_v)$ by using Eq. (\ref{equ:P-GL0})
\STATE $\bm{v} = \frac{\bm{\widehat{v}}}{\|\bm{\widehat{v}}\|_2}$
\STATE $d=\bm{z}^T\bm{v}$
\UNTIL $d$ convergence
\RETURN $\bm{u}$, $\bm{v}$ and $d$
\end{algorithmic}
\end{algorithm}

Note that once the number of elements of every group equals 1 (i.e., $q_i = 1$ for $i =1,2,\cdots,M$), the group $L_0$-norm penalty reduces to $L_0$-norm penalty. Moreover, Algorithm \ref{alg:GL0-SVD} with a small modification can be used to solve SVD($GL_0$, $L_0$), which applies $L_0$-norm as the penalty for the right singular vector $\bm{v}$. In addition, compared to adaptive group lasso \cite{Wang2008A}, we may consider a weighted (adaptive) group $L_0$-penalty. We rewrite $\phi(\bm{z}_u) = [w_1\|\bm{z}^{(1)}\|,\cdots, w_L\|\bm{z}^{(L)}\|]^T$ in Eq. (\ref{equ:P-GL0}), where $w_i$ is a weight coefficient to balance different group-size and it is defined by $w_i = 1/\sqrt{q_i}$, and $q_i$ is the number of elements in group $i$.

\subsection{$\bm{OGL_1}$-SVD}
In some situations, the non-overlapping group structure in group Lasso limits its applicability in practice. For example, a gene can participate in multiple pathways. Several studies have explored the overlapping group Lasso in regression tasks \cite{obozinski2011group,yuan2011efficient}. However, structured sparse SVD with overlapping group structure remains to be solved.

Here we consider the overlapping group situation, where a variable may belong to more than one group. Suppose $\bm{u}$ corresponds to the row-variables of $\bm{X}$ with overlapping groups $\mathcal{G}^u=\{G_1, G_2, \cdots, G_L\}$ and $\bm{v}$ corresponds to the column-variables of $\bm{X}$ with overlapping groups $\mathcal{G}^v=\{G_1, G_2, \cdots, G_M\}$. In other words, $\bm{u}$ and $\bm{v}$ can be respectively divided into $L$ and $M$ groups, which can be represented by $\bm{u}_{G_l}\in \mathbb{R}^{p_l\times 1}, l = 1,...,L$ and $\bm{v}_{G_m}\in \mathbb{R}^{p_m\times 1}, m = 1,...,M$. We define an overlapping group Lasso ($OGL_1$) penalty of $\bm{u}$ as follows \cite{jacob2009group,obozinski2011group,huang2011learning}:
\begin{equation}\label{equ:OGL1-penalty}
  \Omega_{OGL_1}(\bm{u}) = \underset{\mathcal{J} \subseteq \mathcal{G}^u, \mbox{supp}(\phi(\bm{u})) \subseteq \mathcal{J}}{\text{minimize}}~{\sum_{l=1}^L w_l \|\bm{u}_{G_l}\|},
\end{equation}
where $\mbox{supp}(\cdot)$ denotes the index set of non-zero elements for a given vector.

$OGL_1$ is a specific penalty function for structured sparsity. It can lead to the sparse solution, whose supports are unions of predefined overlapping groups of variables. Based on the definition of $OGL_1$, we propose the third group-sparse SVD model as follows:
\begin{equation}
\begin{aligned}\label{equ:L1ogSVD}
&\underset{\bm{u},\bm{v},d}{\text{minimize}} &&\|\bm{X} - d\bm{uv}^T\|_F^2\\
&\text{subject to}                           &&\|\bm{u}\|_2 \leq 1, \Omega_{OGL_1}(\bm{u}) \leq c_u,\\
&                                            &&\|\bm{v}\|_2 \leq 1, \Omega_{OGL_1}(\bm{v}) \leq c_v,
\end{aligned}
\end{equation}
where $c_u$ and $c_v$ are two hyperparameters. We first introduce two latent vectors $\tilde{\bm{u}}$ and $\tilde{\bm{v}}$. Let $\tilde{\bm{u}}^{(l)} = \bm{u}_{G_l}, l = 1,\cdots, L$ and set $\tilde{\bm{u}}= (\tilde{\bm{u}}^{(1)},\cdots,\tilde{\bm{u}}^{(L)})$, which is a column vector with size of $\sum_{l=1}^L |G_l|$. Similarly, we can get $\tilde{\bm{v}}$ based on $\bm{v}$. In addition, we can extend the rows and columns of $\bm{X}$ of $p \times n$ to obtain a new matrix $\tilde{\bm{X}}$ with size of $\sum_{l=1}^L |G_l| \times \sum_{m=1}^M |G_m| $, whose row and column variables are non-overlapping. Thus, solving the problem (\ref{equ:L1ogSVD}) is approximately equivalent to solving a SVD($GL_1$, $GL_1$) for non-overlapping $\tilde{\bm{X}}$. We can obtain an approximate solution of (\ref{equ:L1ogSVD}) by using Algorithm 1. However, if a variable belongs to many different groups, it leads to a large computational burden. For example, given a protein-protein interaction (PPI) network, which contains about 13,000 genes and  250,000 edges. If we consider each edge of the PPI network as a group, then we would construct a high-dimensional matrix $\tilde{\bm{X}}$, which contains 500,0000 rows.

To address this issue, we develop a method based on alternating direction method of multipliers (ADMM) \cite{qin2012structured,boyd2011distributed} to directly solve problem (\ref{equ:L1ogSVD}).
Similar with Eq. (\ref{equ:03}), we first redefine problem (\ref{equ:L1ogSVD}) with its Lagrange form:
\begin{equation}
\begin{aligned}\label{equ:ogl1-lagr}
&\mathcal{L}(\bm{u},\bm{v}) = && -\bm{u}^T\bm{X}\bm{v} + \lambda_1\Omega_{OGL_1}(\bm{u})   \\
&                    && + \lambda_2\Omega_{OGL_1}(\bm{v}) + \eta_1\bm{u}^T\bm{u} + \eta_2\bm{u}^T\bm{v},
\end{aligned}
\end{equation}
where parameters $\lambda_1\geq 0$, $\lambda_2 \geq 0$, $\eta_1\geq 0$ and $\eta_1\geq 0$ are Lagrange multipliers. Inspired by \cite{bolte2014proximal}, we develop an alternating iterative algorithm to minimize it. That is, we optimize the above problem with respect to $\bm{u}$ by fixing $\bm{v}$ and \emph{vice versa}. 
Since $\bm{u}$ and $\bm{v}$ are symmetrical in problem (\ref{equ:ogl1-lagr}), we only need to consider a subproblem with respect to $\bm{u}$ as follows:
\begin{equation}\label{equ:subProblemOfOGL1}
\underset{\bm{u}}{\text{minimize}}~~-\bm{u}^T\bm{z} + \lambda \Omega_{OGL_1}(\bm{u}) + \eta \|\bm{u}\|^2,
\end{equation}
where $\bm{z} = \bm{X}\bm{v}$. Since the overlapping Lasso penalty is a convex function \cite{mosci2010primal}.
We can apply ADMM \cite{qin2012structured,boyd2011distributed} to solve the above problem (\ref{equ:subProblemOfOGL1}).
To obtain the learning algorithm of (\ref{equ:subProblemOfOGL1}), we first introduce an auxiliary $\bm{y}$ and redefine the above problem as follows:
\begin{equation}
\begin{aligned}\label{equ:OGL1po}
& \underset{\bm{u}}{\text{minimize}} &&-\bm{u}^T\bm{z} + \lambda \sum_{l=1}^L w_l\|\bm{y}^{(l)}\|_2 + \eta \|\bm{u}\|^2\\
& \text{subject to}                  &&\bm{y}^{(l)} = \bm{u}_{G_l},~l = 1,\cdots,L.
\end{aligned}
\end{equation}
So the augmented Lagrangian of (\ref{equ:OGL1po}) can be written as follows:
\begin{equation}
\begin{aligned}\label{equ:OGL1_u}
&\mathcal{L_\rho} (\bm{u},\bm{y},\bm{\theta}) = && -\bm{u}^T\bm{z} + \eta \|\bm{u}\|^2 + \sum_{l=1}^L {\bm{\theta}^{(l)}}^T(\bm{y}^{(l)} -\bm{u}_{G_l}) \\
&                    && + \lambda \sum_{l=1}^L w_l\|\bm{y}^{(l)}\|_2 + \frac{\rho}{2}\sum_{l=1}^L \|\bm{y}^{(l)} - \bm{u}_{G_l}\|_2^2,
\end{aligned}
\end{equation}
where Lagrange multipliers $\bm{\theta} = [\bm{\theta}^{(1)};\cdots;\bm{\theta}^{(L)}]$ and $\bm{y} = [\bm{y}^{(1)};\cdots;\bm{y}^{(L)}]$ are two column vectors with $L$ non-overlapping groups.
For convenience, we first define some column-vectors $\tilde{\bm{\theta}}^{(l)}$, $\tilde{\bm{y}}^{(l)}$ and $\tilde{\bm{e}}^{(l)}$ ($l= 1,\cdots,L$), and they have the same size and group structures as $\bm{u}$, where
$\tilde{\bm{\theta}}^{(l)}$ meets that $[\tilde{\bm{\theta}}^{(l)}]_{G_k} = \bm{\theta}^{(l)}$ if $k=l$ and $[\tilde{\bm{\theta}}^{(l)}]_{G_k} = \bm{0}$ otherwise;
$\tilde{\bm{y}}^{(l)}$ meets that $[\tilde{\bm{y}}^{(l)}]_{G_k} = \bm{y}^{(l)}$ if $k=l$ and $[\tilde{\bm{y}}^{(l)}]_{G_k} = \bm{0}$ otherwise;
$\tilde{\bm{e}}^{(l)}$ meets that $[\tilde{\bm{e}}^{(l)}]_{G_k} = \bm{1}$ if $k=l$ and $[\tilde{\bm{e}}^{(l)}]_{G_k} = \bm{0}$ otherwise.
Note that $[\tilde{\bm{\theta}}^{(l)}]_{G_k}$, $[\tilde{\bm{y}}^{(l)}]_{G_k}$ and $[\tilde{\bm{e}}^{(l)}]_{G_k}$ ($k = 1 \cdots L$) respectively represent the elements of $k$-th group of $\tilde{\bm{\theta}}^{(l)}$, $\tilde{\bm{y}}^{(l)}$ and $\tilde{\bm{e}}^{(l)}$. Thus, we have ${\bm{\theta}^{(l)}}^T \bm{u}_{G_l}=\bm{u}^T \tilde{\bm{\theta}}^{(l)}$ and ${\bm{y}^{(l)}}^T \bm{u}_{G_l}=\bm{u}^T \tilde{\bm{y}}^{(l)}$.
So we can obtain the gradient equations with respect to $\bm{u}$ in Eq. (\ref{equ:OGL1_u}) as follows:
\begin{equation}
  \nabla_{\bm{u}}\mathcal{L_\rho} = 2\eta\bm{u} - \bm{z} - \sum_{l=1}^L \tilde{\bm{\theta}}^{(l)} + \rho\left(\sum_{l=1}^L \tilde{\bm{e}}^{(l)}\right)\bullet \bm{u}-\rho \sum_{l=1}^L \tilde{\bm{y}}^{(l)} = \bm{0},
\end{equation}
where ``$\bullet$'' performs element-by-element multiplication. Thus, we can obtain the update rule for $\bm{u}$ and ensure it is a unit vector:
\begin{equation}\label{equ:update-u}
 \bm{u} \leftarrow \frac{\widehat{\bm{u}}}{\|\widehat{\bm{u}}\|},~\mbox{where}~\widehat{\bm{u}} = \bm{z} + \sum_{l=1}^L \tilde{\bm{\theta}}^{(l)} + \rho \sum_{l=1}^L \tilde{\bm{y}}^{(l)}.
\end{equation}
We also obtain the subgradient equations (see \cite{bazaraa2013nonlinear}) with respect to $\bm{y}^{(l)}$ in Eq. (\ref{equ:OGL1_u}) as follows:
\begin{equation}
  \nabla_{\bm{y}^{(l)}}\mathcal{L_\rho} = \lambda w_l \cdot \bm{s}^{(l)} + \bm{\theta}^{(l)} +  \rho (\bm{y}^{(l)} - \bm{u}_{G_l})= \bm{0},
\end{equation}
where $l=1,\cdots,L$, if~$\bm{y}^{(l)} \neq \bm{0}$, then $\bm{s}^{(l)}=\frac{\bm{y}^{(l)}}{\|\bm{y}^{(l)}\|_2}$, otherwise $\bm{s}^{(l)}$ is a vector with $\|\bm{s}^{(l)}\|_2\leq 1$. For convenience, let $\bm{t}^{(l)} = \rho \bm{u}_{G_l}-\bm{\theta}^{(l)}$, we thus develop a block coordinate descent method to learn Lagrange multipliers $\bm{y}$. Since $\bm{y}^{(l)}$ ($l=1,\cdots, L$) are independent. Thus,  $\bm{y}^{(l)}$ ($l=1,\cdots, L$) can be updated in parallel according to the following formula:
\begin{eqnarray}\label{equ:update-y}
  \bm{y}^{(l)}\leftarrow
  \begin{cases}
    \frac{1}{\rho}\left(1- \frac{\lambda w_l}{\|\bm{t}^{(l)}\|_2}\right)\bm{t}^{(l)}, &\mbox{if}~\|\bm{t}^{(l)}\|_2 > \lambda w_l,\cr
    \bm{0}, &\mbox{otherwise}.
  \end{cases}
\end{eqnarray}
Based on ADMM \cite{boyd2011distributed}, we also obtain the update rule for $\bm{\theta}$ as follows:
\begin{equation} \label{equ:update-theta}
 \bm{\theta}^{(l)} \leftarrow \bm{\theta}^{(l)} + \rho (\bm{y}^{(l)} -\bm{u}_{G_l}),~~l = 1,\cdots, L.
\end{equation}
Combining Eqs. (\ref{equ:update-u}), (\ref{equ:update-y}) and (\ref{equ:update-theta}), we thus get an ADMM based method to solve problem (\ref{equ:OGL1po}) (Algorithm 3). Note that the output of Algorithm 3 is a set of selected group indexes, defined as $T$. For example, if $\bm{y} = [\bm{y}^{(1)};\bm{y}^{(2)};\bm{y}^{(3)}]$, $\bm{y}^{(1)}=\bm{0}$, $\bm{y}^{(2)} \neq \bm{0}$, and $\bm{y}^{(3)} \neq \bm{0}$, then $T=\{2,3\}$.
\begin{algorithm}[h]
\caption{ADMM method for problem (\ref{equ:OGL1po})} \label{alg:L0-OGPCA}
\begin{algorithmic}[1]
\REQUIRE $\bm{z}\in \mathbb{R}^{p}$, $\mathcal{G}$, $\lambda$, $\rho > 0$
\STATE Initialize $\bm{\theta}$ and $\bm{y}$
\REPEAT
\STATE Updating $\bm{u}$ with fixed $\bm{y}$ and $\bm{\theta}$ using Eq. (\ref{equ:update-u})
\STATE Updating $\bm{y}$ with fixed $\bm{u}$ and $\bm{\theta}$ using Eq. (\ref{equ:update-y})
\STATE Updating $\bm{\theta}$ with fixed $\bm{u}$ and $\bm{y}$ using Eq. (\ref{equ:update-theta})
\UNTIL convergence
\STATE $T = \{g: \|\bm{y}^{(g)}\|_2>0, g \in \{1,\cdots, |\mathcal{G}|\}\}$
\RETURN $T$
\end{algorithmic}
\end{algorithm}
\begin{algorithm}[h]
\caption{$OGL_1$-SVD or SVD($OGL_1$, $OGL_1$)} \label{alg:OGL1-SVD}
\begin{algorithmic}[1]
\REQUIRE Matrix $\bm{X}\in \mathbb{R}^{p\times n}$, $\lambda_u$, and $\lambda_v$; $\mathcal{G}^u$ and $\mathcal{G}^v$
\ENSURE $\bm{u}$, $\bm{v}$ and $d$
\STATE Initialize $\bm{v}$ with $\|\bm{v}\|=1$
\REPEAT
\STATE Let $\bm{z}_u=\bm{Xv}$
\STATE Get the active groups $T_u$ by Algorithm 3 with $\bm{z}_u$ and $\mathcal{G}^u$ as the input
\STATE $\widehat{\bm{u}}=\bm{z}_u \circ 1_{T_u}$
\STATE $\bm{u} = \frac{\widehat{\bm{u}}}{\|\widehat{\bm{u}}\|}$
\STATE Let $\bm{z}_v=\bm{X}^T\bm{u}$
\STATE Get the active groups $T_v$ by Algorithm 3 with $\bm{z}_v$ and $\mathcal{G}^v$ as the input
\STATE $\widehat{\bm{v}}=\bm{z}_v \circ 1_{T_v}$
\STATE $\bm{v} = \frac{\widehat{\bm{v}}}{\|\widehat{\bm{v}}\|}$
\STATE $d=\bm{z}^T\bm{v}$
\UNTIL $d$ convergence
\RETURN $\bm{u}$, $\bm{v}$ and $d$
\end{algorithmic}
\end{algorithm}

In summary, based on the ADMM algorithm (Algorithm 3), we adopt an alternating iterative strategy (Algorithm 4) to solve SVD($OGL_1$,$OGL_1$). In Algorithm 4, the operation $\bm{x}=\bm{z} \circ 1_{T}$ denotes if group $l\in T$, then $\bm{x}_{G_l} = \bm{z}_{G_l}$, and the remaining elements of $\bm{x}$ are zero.

\subsection{$\bm{OGL_0}$-SVD}
Here we define an overlapping group $L_0$-norm penalty ($OGL_0$) of $\bm{u}$ as follows:
\begin{equation}
  \Omega_{OGL_0}(\bm{u}) = \underset{\mathcal{J} \subseteq \mathcal{G}^u,~\mbox{supp}(\phi(\bm{u})) \subseteq \mathcal{J}}{\text{minimize}}~{\sum_{l=1}^L \mathds{1} (\|\bm{u}_{G_l}\|\neq0)},
\end{equation}
where $\mbox{supp}(\cdot)$ denotes the index set of non-zero elements for a given vector.

Based on the definition of $OGL_0$, we propose the fourth group-sparse SVD model with overlapping group $L_0$-norm penalty ($OGL_0$-SVD) as follows:
\begin{equation}
\begin{aligned}\label{equ:L0-OGSVD}
& \underset{\bm{u},\bm{v},d}{\text{minimize}} && \|\bm{X} - d\bm{uv}^T\|_F^2\\
& \text{subject to}                         && \|\bm{u}\|_2 \leq 1, \Omega_{OGL_0}(\bm{u}) \leq k_u,\\
&                                           && \|\bm{v}\|_2 \leq 1, \Omega_{OGL_0}(\bm{v}) \leq k_v.
\end{aligned}
\end{equation}
Similarly, we solve the above problem by using an alternating iterative method. Fix $\bm{u}$ (or $\bm{v}$), we transform the original optimization problem into a projection problem with overlapping group $L_0$-norm penalty.

Fix $\bm{v}$ in problem (\ref{equ:L0-OGSVD}) and let $\bm{z}= \bm{X}\bm{v}$, thus the problem can be written into a projection problem with overlapping group $L_0$-norm penalty:
\begin{equation}\label{equ:ogsp}
  \underset{\|\bm{u}\| \leq 1}{\text{minimize}}~-\bm{z}^T\bm{u},~~\mbox{s.t.}~\Omega_{OGL_0}(\bm{u}) \leq k_u.
\end{equation}
To solve the above problem, we introduce $\bm{y}$ and obtain the above problem in a new way:
\begin{equation}\label{equ:ogsp-2}
  \underset{\|\bm{u}\| \leq 1, \bm{y}}{\text{minimize}}~-\bm{z}^T\bm{u},~~\mbox{s.t.}~\Omega_{GL_0}(\bm{y}) \leq k_u,~\bm{y}^{(l)} = \bm{u}_{G_l},
\end{equation}
where $l = 1,\cdots,L$ and $\bm{y} = [\bm{y}^{(1)};\cdots;\bm{y}^{(L)}]$.

The above problem contains overlapping group-sparse induced penalty with $L_0$-norm. Thus, it is difficult to solve the exact solution of problem (\ref{equ:ogsp-2}). To this end, we use an approximate method, which replaces $\bm{z}^T\bm{u}$ by using $\sum_l \bm{z}_{G_l}^T\bm{u}_{G_l}$. Since $\bm{y}^{(l)} = \bm{u}_{G_l}$ in problem (\ref{equ:ogsp-2}), we have $\sum_l \bm{z}_{G_l}^T\bm{u}_{G_l} = \sum_l \bm{z}_{G_l}^T\bm{y}^{(l)}$. Thus, problem (\ref{equ:ogsp-2}) approximately reduces to the below problem,
\begin{equation}\label{equ:ogsp-3}
  \underset{\|\bm{u}\| \leq 1, \bm{y}}{\text{minimize}}~-\sum_l \bm{z}_{G_l}^T\bm{y}^{(l)},~~\mbox{s.t.}~\Omega_{GL_0}(\bm{y}) \leq k_u,~\bm{y}^{(l)} = \bm{u}_{G_l}.
\end{equation}
Since $\bm{y}$ contains a non-overlapping structure, we can easily get the optimal solution of the above problem on $\bm{u}$ and $\bm{y}$.
To sum up, we obtain an approximate solution of (\ref{equ:ogsp}) as Theorem 2 suggests.
\begin{theorem}
The approximate solution of (\ref{equ:ogsp}) is $\frac{\widehat{\mathcal{P}}_{OGL_0}(\bm{z})}{\|\widehat{\mathcal{P}}_{OGL_0}(\bm{z})\|_2}$ and
\begin{eqnarray}\label{equ:theorem2}
  [\widehat{\mathcal{P}}_{OGL_0}(\bm{z})]_{G_i}=
  \begin{cases}
    \bm{z}_{G_i}, &\mbox{if}~i \in \mbox{supp}(\phi(\bm{z}),k_u),\cr
    \bm{0}, &\mbox{otherwise},
  \end{cases}
\end{eqnarray}
where $i = 1,2,\cdots,L$, $\phi(\bm{z})=[\|\bm{z}_{G_1}\|;\cdots;\|\bm{z}_{G_L}\|]$ and $\mbox{supp}(\phi(\bm{z}), k_u)$ denotes the set of indexes of the largest $k_u$ elements of $\phi(\bm{z})$.
\end{theorem}
Briefly, Theorem 2 shows that approximately solving the problem (\ref{equ:ogsp}) is equivalent to keeping elements of $k_u$ groups with the largest $k_u$ group-norm values and the other elements are zeros. 

Fix $\bm{u}$ in problem (\ref{equ:L0-OGSVD}) and let $\bm{z}= \bm{X}^T\bm{u}$, thus the problem (\ref{equ:L0-OGSVD}) reduces to the following one:
\begin{equation}\label{equ:ogsp-v}
  \underset{\|\bm{v}\| \leq 1}{\text{minimize}}~-\bm{z}^T\bm{v},~~\mbox{s.t.}~\Omega_{OGL_0}(\bm{v}) \leq k_v.
\end{equation}
Similarly, based on Theorem 2, we can obtain the approximate solution of (\ref{equ:ogsp-v}) as $\frac{\widehat{\mathcal{P}}_{OGL_0}(\bm{z})}{\|\widehat{\mathcal{P}}_{OGL_0}(\bm{z})\|_2}$.
Finally, we propose an alternating iterative method based on an approximate method to solve problem (\ref{equ:L0-OGSVD}) (Algorithm \ref{alg:l0ogsvd}).
\begin{algorithm}[h]
\caption{$OGL_0$-SVD or SVD($OGL_0$, $OGL_0$)} \label{alg:l0ogsvd}
\begin{algorithmic}[1]
\REQUIRE Matrix $\bm{X}\in \mathbb{R}^{p\times n}$, $k_u$ and $k_v$; $\mathcal{G}^u$ and $\mathcal{G}^u$
\ENSURE $\bm{u}$, $\bm{v}$ and $d$
\STATE Initialize $\bm{v}$ with $\|\bm{v}\|_2=1$
\REPEAT
\STATE Let $\bm{z}_u=\bm{Xv}$
\STATE $\bm{\widehat{u}} = \widehat{\mathcal{P}}_{OGL_0}(\bm{z}_u)$ by using Eq. (\ref{equ:theorem2})
\STATE $\bm{u} = \frac{\bm{\widehat{u}}}{\|\bm{\widehat{u}}\|_2}$
\STATE Let $\bm{z}_v=\bm{X}^T\bm{u}$
\STATE $\bm{\widehat{v}} = \widehat{\mathcal{P}}_{OGL_0}(\bm{z}_v)$ by using Eq. (\ref{equ:theorem2})
\STATE $\bm{v} = \frac{\bm{\widehat{v}}}{\|\bm{\widehat{v}}\|_2}$
\STATE $d=\bm{z}^T\bm{v}$
\UNTIL $d$ convergence
\RETURN $\bm{u}$, $\bm{v}$ and $d$
\end{algorithmic}
\end{algorithm}

\subsection{Convergence analysis}
Inspired by \cite{Tseng2001Convergence,grippo2000convergence}, for a two-block coordinate problem, if its objective function of each subproblem is strictly convex, then there exists a unique global optimal solution for this problem. The Gauss-Seidel method can effectively solve such a two-block coordinate problem and it converges to a critical point for any given initial (see \cite{razaviyayn2013unified} and the references therein). We note both the proposed $GL_1$-SVD (Algorithm 1) and $OGL_1$-SVD (Algorithm 4) are Gauss-Seidel type of methods and those subproblems of $GL_1$-SVD and $OGL_1$-SVD models are strictly convex. Thus, $GL_1$-SVD and $OGL_1$-SVD algorithms are convergent.

Next we discuss the convergence of $GL_0$-SVD (Algorithm 2). In \cite{bolte2014proximal}, the authors developed a class of methods based on a proximal gradient strategy to solve a broad class of nonconvex and nonsmooth problems:
\begin{equation}\label{ALM}
  \underset{\bm{u},~\bm{v}}{\text{minimize}}~F(\bm{u},\bm{v}) = f(\bm{u})+ g(\bm{v}) + H(\bm{u},\bm{v}),
\end{equation}
where $f(\bm{u})$ and $g(\bm{v})$ are nonconvex and nonsmooth functions and $H(\bm{u},\bm{v})$ is a smooth function (also see \cite{pock2016inertial,yang2017proximal,xu2017globally}). $GL_0$-SVD model can be seen as such a problem:
\begin{subequations}
\begin{align}
H(\bm{u},\bm{v}) & = -\bm{u}^T\bm{Xv},\\
f(\bm{u}) & =
  \begin{cases}
    0, &\mbox{if}~\|\bm{u}\|=1,~\Omega_{OGL_0}(\bm{u})\leq k_u,\cr
    +\infty, &\mbox{otherwise}.
  \end{cases} \\
g(\bm{v}) & =
  \begin{cases}
    0, &\mbox{if}~\|\bm{v}\|=1,~\Omega_{OGL_0}(\bm{v})\leq k_v,\cr
    +\infty, &\mbox{otherwise}.
  \end{cases}
\end{align}
\end{subequations}
Note $F(\bm{u},\bm{v}) = f(\bm{u})+ g(\bm{v}) + H(\bm{u},\bm{v})$ of $GL_0$-SVD is semialgebraic and meets the KL property (Regarding the semialgebraic and KL property, please see \cite{bolte2014proximal,pock2016inertial,xu2017globally}). Based on the Theorem 1 in \cite{bolte2014proximal} (also see Theorem 2 in \cite{xu2017globally}), we can obtain that $GL_0$-SVD algorithm converges to a critical point.

In a word, $GL_1$-SVD (Algorithm 1), $GL_0$-SVD (Algorithm 2) and $OGL_1$-SVD (Algorithm 4) converge to their corresponding critical points. Although $OGL_0$-SVD (Algorithm 5) applies an approximate strategy, it has a good convergence in practice.

\subsection{Group-sparse SVD for edge-guided gene selection}
Given a high-dimensional data (e.g., gene expression data) and a prior network (e.g., a gene interaction network), we can consider a special edge group structure, in which each edge (e.g., a gene interaction) is considered as a group. In our study, the gene interaction network is regarded as the graph $(V, E)$ where $V = \{1,\cdots,p\}$ is the set of nodes (genes) and $E$ is the set of edges (gene interactions).
SVD($OGL_0$, $L_0$) can be applied to analyze such high-dimensional gene expression data via integrating group information $\mathcal{G}^u = E$. The estimated sparse solution is used for gene selection.

\subsection{Learning multiple factors}
To identify the next gene module, we subtract the signal of current pair of singular vectors from the input data (i.e., $\bm{X}:=\bm{X}-d\bm{u}^T\bm{v}$), and then apply SVD($OGL_0$,$L_0$) again to identify the next pair of sparse singular vectors. Repeat this step for $r$ times, we obtained $r$ pairs of sparse singular vectors and get a rank $r$ approximation of matrix $\bm{X}$.

\section{Simulation Study}
\begin{figure*}[htbp]
  \centering
  \includegraphics[width=1.01\linewidth]{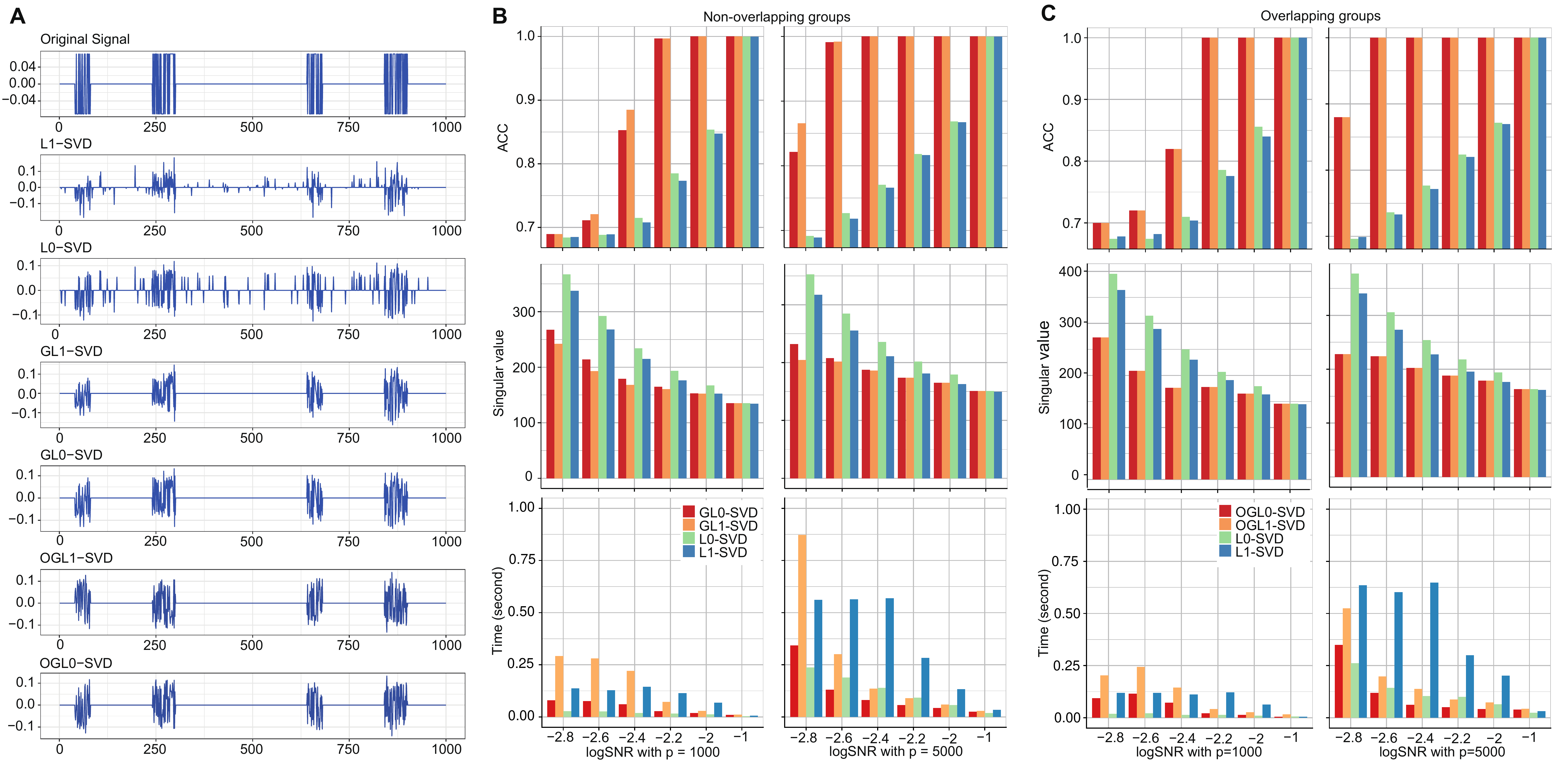}
  \caption{Evaluation results of $GL_1$-SVD, $GL_0$-SVD, $OGL_1$-SVD and $OGL_0$-SVD and comparison with $L_1$-SVD and $L_0$-SVD using the simulated data $\bm{X}$ with prior (non-) overlapping group information. (A) Illustration of all methods in the simulated $\bm{X}$ with $p=1000$ and $\mbox{logSNR} = -2$.
  (B) and (C) Results of all methods in terms of ACC, singular value $d$ and time (second) for cases with GR and OGR respectively. Each bar entry is the average over 50 replications. The input $\bm{X}s\in \mathbb{R}^{p\times 100}$ vary with different $p$=1000 and 5000, and $\mbox{logSNR}$=-2.8, -2.6, -2.4, -2.2, -2 and -1, respectively.   }\label{fig02}
\end{figure*}

In the section, we applied these group sparse SVD methods ($GL_1$-SVD, $GL_0$-SVD, $OGL_1$-SVD and $OGL_0$-SVD) to a set of simulated data and compared their performance with several sparse SVD method without using prior group information including $L_0$-SVD \cite{Asteris2016,min2015novel}, $L_1$-SVD \cite{lee2010biclustering}. We generated two types of simulation data with respect to non-overlapping group structure (GR) and overlapping group structure (OGR) respectively (Fig. 1A).

Without loss of generality, we first generated $\bm{u}$ and $\bm{v}$ for GR and OGR cases, and then generated a rank-one data matrix by using formula
\begin{equation}\label{equ:getX}
\bm{X} = d\bm{u}\bm{v}^T + \gamma \bm{\epsilon},
\end{equation}
where $d=1$, $\bm{\epsilon}_{ij} \stackrel{\text{i.i.d.}}{\sim} \mathcal{N}(0,1)$ and $\gamma$ is a nonnegative parameter to control the signal-to-noise ratio (SNR). The logarithm of SNR (logSNR) is defined by:
\begin{equation}\label{equ:logSNR}  
  \mbox{logSNR} = \log_{10}\bigg( \frac{\|d\bm{u}\bm{v}\|_F^2}{\mathbf{E}(\|\gamma \bm{\epsilon}\|_F^2)} \bigg) = \log_{10}\bigg(\frac{\|d\bm{u}\bm{v}\|_F^2}{\gamma^2np}\bigg),
\end{equation}
where $\mathbf{E}(\|\gamma \bm{\epsilon}\|_F^2)$ denotes the expected sum of squares of noise.

We evaluated the performance of all methods by the following measures including true positive rate (TPR), true negative rate (TNR), false positive rate (FPR), false discovery rate (FDR) and accuracy (ACC). They are defined as follows:
$$ \mbox{TPR} = \frac{\mbox{TP}}{\mbox{P}},~\mbox{TNR} = \frac{\mbox{TN}}{\mbox{N}},~\mbox{FPR} = \frac{\mbox{FP}}{\mbox{N}},$$
$$ \mbox{FDR} = \frac{\mbox{FP}}{\mbox{TP+FP}},~\mbox{ACC} = \frac{\mbox{TP+TN}}{\mbox{TP+FP+FN+TN}},$$
where \mbox{P} denotes the number of positive samples, \mbox{N} denotes the number of negative samples, \mbox{TP} denotes the number of true positive, \mbox{TN} denotes the number of true negative, \mbox{FP} denotes the number of false positive, and \mbox{FN} denotes the number of false negative, respectively.

\subsection{Non-overlapping group structure (GR)}
We generated the simulated data matrix $\bm{X} \in \mathbb{R}^{p\times n}$ with $n = 100$ samples without groups, and $p$ row variables with 50 groups.
We first generated $\bm{v} = \mbox{rnorm}(n, \mbox{mean} = 0, \mbox{sd} = 1)$, which samples $n$ elements from the standard normal distribution. Then we generated $\bm{u} = [ \bm{u}_{G_1}; \bm{u}_{G_2};\cdots;\bm{u}_{G_{50}}] $ with 50 groups where if $i \in \{3,4,13,14,15,33,34,43,44,45\}$, and $\bm{u}_{G_i} = \mbox{sample}(\{-1,1\}, q)$, which samples $q$ elements from $\{-1,1\}$, and $q$ denotes the number of members of a group, otherwise $\bm{u}_{G_i} = \bm{0}$. Finally, we obtained the simulated matrix $\bm{X}$ by using Eq.~(\ref{equ:getX}).

Here we first considered $q \in \{20,100\}$ and $\mbox{logSNR} \in \{-1,-2,-2.2,-2.4,-2.6,-2.8\}$ to generate simulated data with GR. Given a $\mbox{logSNR}$, suppose $\bm{u}$, $\bm{v}$ and $d$ are known, then we got a $\gamma$ by using Eq.~(\ref{equ:logSNR}). For each pair ($q$, $\mbox{logSNR}$), we generated 50 simulated matrices $\bm{X}s$.

Here, we evaluated the performance of $GL_1$-SVD and $GL_0$-SVD with $k_u = 10$ (groups) in this simulated data. For comparison, we forced the identified $\bm{u}$ to contain 200 non-zero elements if $q = 20$ (i.e., the number of rows of $\bm{X}$ is $p=1000$) and 1000 non-zero elements if $q=100$ (i.e., $p=5000$) for $L_1$-SVD and $L_0$-SVD by tuning their parameters. For visualization, we first tested $GL_1$-SVD and $GL_0$-SVD with $k_v = 10$ to a GR simulated $\bm{X}$ with $p=1000$ and $\mbox{logSNR} = -2$, to explain their performance and compared them with other methods (Fig. 2A).
Obviously $GL_1$-SVD and $GL_0$-SVD can improve performance of variable selection by integrating non-overlapping group information of variables. Further, we tested our methods on more GR simulation data (Fig. 2B and TABLE 2). We also found that the performance of $GL_0$-SVD and $GL_1$-SVD are significantly superior to that of $L_0$-SVD, $L_1$-SVD in terms of different logSNRs with $p = 1000$ or $p = 5000$ (Fig.~2B). In particular, the greater the noise of simulated data, the better the performance of our methods. Furthermore, compared with $GL_1$-SVD, $GL_0$-SVD obtains higher singular values for different logSNRs (Fig.~2B). We also compare the different algorithms on GPU time of an ordinary personal computer, all the algorithm takes less than one second. Computational results illustrate that our models can enhance the power of variable selection by integrating group information of variables.

\subsection{Overlapping group structure (OGR)}
To generate OGR simulated data, we first generated $\bm{v}$ with $n = 100$ (samples) and its elements are from a standard normal distribution, i.e., $\bm{v} = \mbox{rnorm}(n, \mbox{mean} = 0, \mbox{sd} = 1)$. Then we generated $\bm{u}$ with overlapping groups. We considered an overlapping group structure for $\bm{u}$ as follows: $G_1 = \{1, 2,..., 2t\}$, $G_2 = \{t+1, t+2,...,3t\}$, $\cdots$, $G_{48} = \{47t+1,47t+2,...,49t\}$, $G_{49} = \{48t+1,48t+2,...,50t\}$, where every group and its adjacent groups overlap half of the elements. Note that the dimension of $\bm{u}$ is $p = 50t$. If $i \in \{3, 13,14, 33, 43,44\}$ (active groups), then $\bm{u}_{G_i} = \mbox{sample}(\{-1,1\}, 2t)$ ($2t$ denotes the number of members of a group), otherwise $\bm{u}_{G_i} = \bm{0}$. We considered $t \in \{20,100\}$ and $\mbox{logSNR} \in \{-1,-2,-2.2,-2.4,-2.6,-2.8\}$ for generating OGR simulated data. Note that once $\bm{u}$, $\bm{v}$ and $\mbox{logSNR}$ is given, we could generate the simulated matrix $\bm{X} \in \mathbb{R}^{p\times n}$ using Eq. (\ref{equ:getX}) where $d=1$. For each pair ($q$, $\mbox{logSNR}$), we generated 50 simulated matrices $\bm{X}$s.

For visualization, we first applied $OGL_1$-SVD and $OGL_0$-SVD with $k_v = 5$ onto a simulated $\bm{X}$ with OGR, $p=1000$ and $\mbox{logSNR} = -2$, to explain their performance and compared them with other methods (Fig. 2A). Obviously $OGL_1$-SVD and $OGL_0$-SVD can improve performance of variable selection by integrating overlapping group information of variables. Further, we tested our methods on more OGR simulation data (Fig. 2B and TABLE 2). For comparison, we forced the identified $\bm{u}$ to contain 200 non-zero elements if $t = 20$ (i.e., the number of rows of $\bm{X}$ is $p=1000$) and 1000 non-zero elements $t=100$ for $L_1$-SVD and $L_0$-SVD by tuning their parameters. The performance of $OGL_0$-SVD and $OGL_1$-SVD are significantly superior to that of $L_0$-SVD, $L_1$-SVD in terms of different logSNRs with $p = 1000$ or $p = 5000$ (Fig.~2B). $OGL_1$-SVD and $OGL_0$-SVD get similar singular values, whereas $OGL_1$-SVD needs more time in terms of different logSNRs with $p = 1000$ or $p = 5000$ (Fig.~2B).

Finally, we also investigated the effect of our methods with different sizes of data $\bm{X}s$. In the simulated data, there are 10 active groups (each group contains $q$ members) for GR cases, and there are 5 active groups for OGR cases and each group contains $2t$ members. We set $q = t$, thus we can set a common original signal of $\bm{u}$ for GR and OGR cases. Based on the definition of $\bm{u}$ and $\bm{v}$, we set $q \in \{20,40,100,160,200\}$ (i.e., $p \in \{1000, 2000, 5000, 8000, 10000\}$) and logSNR=-2.8 to generate $\bm{X}$ using Eq. (\ref{equ:getX}). For each pair ($q$, $\mbox{logSNR}$), we generated 50 simulated matrices $\bm{X}$s. We applied $GL_0$-SVD, $GL_1$-SVD, $L_0$-SVD and $L_1$-SVD onto the simulated data and compared their performance in different ways (TABLE 2). In summary, the group-sparse SVD methods obtain higher TPR, TNR and ACC (lower FPR and FDR) than $L_1$-SVD and $L_0$-SVD do (TABLE 2). Naturally, the group-sparse methods spent a bit more time, and obtain lower singular value $d$.

\begin{table}[htbp]
\caption{Evaluation results of $GL_0$-SVD, $GL_1$-SVD, $L_0$-SVD and $L_1$-SVD in terms of TPR, TNR, FPR, FDR, ACC, singular value $d$ and time (second). The input $\bm{X}\in \mathbb{R}^{p\times 100}$ vary with different $p$=1000, 2000, 5000, 8000 and 10000. All these simulated $\bm{X}s$ are generated at $\mbox{logSNR}=-2.8$. Each entry is an average over 50 replications.}
\centering
\resizebox{\columnwidth}{!}{
\begin{tabular}{|c|c|c|c|c|c|c|}
\hline
\multicolumn{1}{|c}{} &\multicolumn{6}{c|}{\textbf{$\bm{p}$ = 1000}}\\
\hline
  & $L_1$-SVD & $L_0$-SVD & $GL_1$-SVD & $GL_0$-SVD & $OGL_1$-SVD & $OGL_0$-SVD \\
\hline
  TPR & 0.21 & 0.21 & 0.22 & 0.22 & 0.23 & 0.23 \\
  TNR & 0.80 & 0.80 & 0.81 & 0.81 & 0.82 & 0.82 \\
  FPR & 0.20 & 0.20 & 0.19 & 0.19 & 0.18 & 0.18 \\
  FDR & 0.79 & 0.79 & 0.78 & 0.78 & 0.76 & 0.76 \\
  ACC & 0.68 & 0.68 & 0.69 & 0.69 & 0.70 & 0.70 \\
  d & 337.91 & 367.61 & 242.38 & 267.73 & 257.94 & 258.08 \\
  time & 0.10 & 0.02 & 0.27 & 0.07 & 0.25 & 0.12 \\
\hline
\multicolumn{1}{|c}{} &\multicolumn{6}{c|}{\textbf{$\bm{p}$ = 2000}}\\
\hline
  TPR & 0.21 & 0.21 & 0.25 & 0.24 & 0.23 & 0.23 \\
  TNR & 0.80 & 0.80 & 0.81 & 0.81 & 0.82 & 0.82 \\
  FPR & 0.20 & 0.20 & 0.19 & 0.19 & 0.18 & 0.18 \\
  FDR & 0.79 & 0.79 & 0.75 & 0.76 & 0.75 & 0.75 \\
  ACC & 0.68 & 0.68 & 0.70 & 0.70 & 0.70 & 0.70 \\
  d & 436.70 & 480.66 & 284.84 & 329.42 & 317.45 & 317.65 \\
  time & 0.20 & 0.05 & 0.40 & 0.14 & 0.36 & 0.18 \\
\hline
\multicolumn{1}{|c}{} &\multicolumn{6}{c|}{\textbf{$\bm{p}$ = 5000}}\\
\hline
  TPR & 0.22 & 0.23 & 0.66 & 0.55 & 0.55 & 0.55 \\
  TNR & 0.81 & 0.81 & 0.92 & 0.89 & 0.90 & 0.90 \\
  FPR & 0.19 & 0.19 & 0.08 & 0.11 & 0.10 & 0.10 \\
  FDR & 0.78 & 0.77 & 0.34 & 0.45 & 0.40 & 0.40 \\
  ACC & 0.69 & 0.69 & 0.87 & 0.82 & 0.83 & 0.83 \\
  d & 634.84 & 705.79 & 409.22 & 464.14 & 447.67 & 447.21 \\
  time & 0.48 & 0.21 & 0.80 & 0.29 & 0.62 & 0.34 \\
\hline
\multicolumn{1}{|c}{} &\multicolumn{6}{c|}{\textbf{$\bm{p}$ = 8000}}\\
\hline
  TPR & 0.24 & 0.25 & 0.95 & 0.91 & 0.83 & 0.83 \\
  TNR & 0.81 & 0.81 & 0.99 & 0.98 & 0.97 & 0.97 \\
  FPR & 0.19 & 0.19 & 0.01 & 0.02 & 0.03 & 0.03 \\
  FDR & 0.76 & 0.75 & 0.05 & 0.09 & 0.13 & 0.13 \\
  ACC & 0.70 & 0.70 & 0.98 & 0.97 & 0.94 & 0.94 \\
  d & 778.10 & 869.29 & 555.55 & 588.70 & 569.39 & 569.11 \\
  time & 0.89 & 0.44 & 0.79 & 0.31 & 0.51 & 0.30 \\
\hline
\multicolumn{1}{|c}{} &\multicolumn{6}{c|}{\textbf{$\bm{p}$ = 10000}}\\
\hline
  TPR & 0.25 & 0.27 & 1.00 & 0.99 & 0.92 & 0.92 \\
  TNR & 0.81 & 0.82 & 1.00 & 1.00 & 0.98 & 0.98 \\
  FPR & 0.19 & 0.18 & 0.00 & 0.00 & 0.02 & 0.02 \\
  FDR & 0.75 & 0.73 & 0.00 & 0.01 & 0.07 & 0.06 \\
  ACC & 0.70 & 0.71 & 1.00 & 1.00 & 0.97 & 0.97 \\
  d & 863.06 & 964.79 & 634.32 & 655.52 & 645.11 & 645.85 \\
  time & 1.25 & 0.60 & 0.80 & 0.32 & 0.48 & 0.31 \\
\hline
\end{tabular}
}
\end{table}

\begin{figure*}[t]
  \centering
  \includegraphics[width=1.00\linewidth]{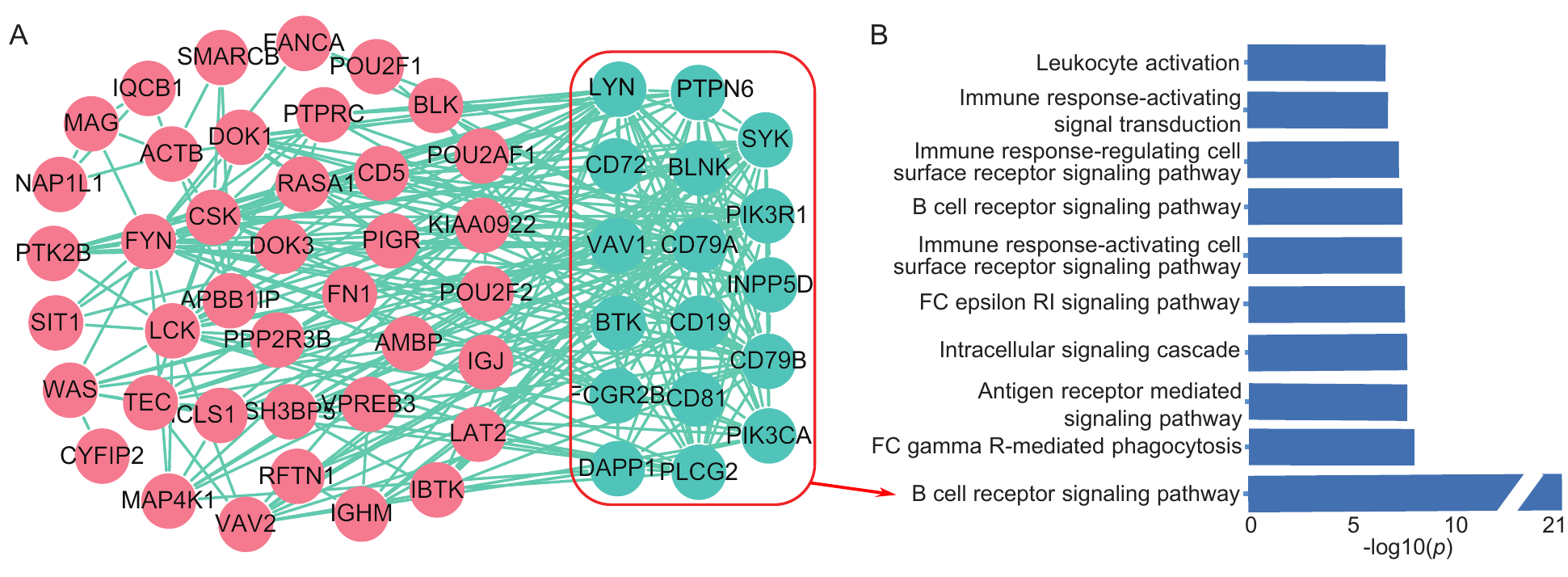}
  \caption{Results of ``CGP + PPI'' dataset. (A) The gene set in the identified module by SVD($OGL_0$, $L_0$) corresponds to a subnetwork of the prior PPI network. The gene subnetwork contain 56 genes and 18 of them (the circled ones) belong to a KEGG pathway, \emph{B~cell~receptor signaling pathway}. (B) Top 10 enriched KEGG and GOBP pathways of this module are shown. Enrichment score was computed by -log10($p$) ($p$ is Benjamini-Hochberg adjusted).}
\end{figure*}

\section{Biological Applications}

We applied our models to two gene expression data of two well-known large-scale projects.
\subsection{Biological data}\label{Data}
\textbf{CGP expression data.} We first downloaded a gene expression dataset from the Cancer Genome Project (CGP) \cite{garnett2012systematic} with 13321 genes across 641 cell lines (samples). The 641 cell lines are derived from different tissues and cancer types.

\textbf{TCGA expression data.} We also obtained twelve cancer gene expression datasets for twelve cancer types across about 4000 cancer samples (\url{http://www.cs.utoronto.ca/~yueli/PanMiRa.html}), which is downloaded from TCGA database (\url{http://cancergenome.nih.gov/}). The twelve cancer types consist of Bladder urothelial carcinoma (BLCA, 134 samples), Breast invasive carcinoma (BRCA, 847 samples), Colon and rectum carcinoma (CRC, 550 samples), Head and neck squamous-cell carcinoma (HNSC, 303 samples), Kidney renal clear-cell carcinoma  (KIRC, 474 samples), Brain lower grade glioma (LGG, 179 samples), Lung adenocarcinoma (LUAD, 350 samples), Lung squamous-cell carcinoma (LUSC, 315 samples), Prostate adenocarcinoma (PRAD, 170 samples), Skin cutaneous melanoma (SKCM, 234 samples), Thyroid carcinoma (THCA, 224 samples), Uterine corpus endometrioid carcinoma (UCEC, 478 samples). We normalized each gene expression dateset of a given cancer type using R function \emph{scale}. Furthermore, we also downloaded the corresponding clinical data of the above 12 cancer types from Firehose (\url{http://firebrowse.org/}).

\textbf{KEGG pathway data.} To integrate the pathway group information with SVD($GL_0$, $L_0$), we also downloaded the KEGG \cite{kanehisa2000kegg} gene pathways from the Molecular Signatures Database (MSigDB) \cite{subramanian2005gene}. We considered a KEGG pathway as a group and removed all the KEGG pathways with $>100$ genes. Finally, we obtained 151 KEGG pathways across 2778 genes by only considering the intersection genes between the CGP gene expression and KEGG pathways data. On average, a gene pathway contains about 40 genes.

\textbf{PPI network.} We also downloaded a protein-protein interaction (PPI) network from Pathway Commons (\url{http://www.pathwaycommons.org/}). In our application, we also considered the edge set of the PPI network as overlapping groups $\mathcal{G}^u$, i.e, a gene (protein) interaction edge represents a group.

\textbf{Data collection.} Finally, we generated three biological datasets to assess our models:
\begin{itemize}
  \item Dataset 1: ``CGP + PPI''. This dataset was obtained by combining CGP gene expression and PPI network data with 13,321 genes, 641 samples and 262,462 interactions.
  \item Dataset 2: ``TCGA + PPI''. This dataset was obtained by combining TCGA gene expression and PPI network data. For each TCGA cancer type, we obtain the expression of 10,399 genes and a PPI network of 10399 genes and 257039 interactions.
  \item Dataset 3: ``CGP + KEGG''. This dataset was obtained by combining CGP gene expression and KEGG pathway data including gene expression of 641 samples and 151 KEGG pathways across 2778 genes.
\end{itemize}

\textbf{Biological analysis of gene modules:} To assess whether the identified modules have significant biological functions, we employed the bioinformatics tool DAVID (https://david.ncifcrf.gov/) \cite{Huang2009Systematic} to perform gene enrichment analysis with GO biological processes (BP) and KEGG pathways. The terms with Benjamin corrected $p$-value $<$ 0.05 are considered as significant ones.

For each cancer type, we first used SVD($OGL_0$, $L_0$) to find a gene module with similar expressions (Fig. 1). To analyze the clinical relevance of each module in a given cancer type, we ran a multivariate multivariate Cox proportional hazard model to obtain the prognostic scores for the patients of this cancer type. It was implemented by using \emph{predict} function in the R package `survival' with type = ``lp''. Then we divided the patients of this cancer type into low-risk and high-risk groups based on the prognostic scores. Finally, we assessed the overall survival difference between the two groups using log-rank test and drew a Kaplan-Meier (KM) curve for visualization.

\subsection{Application to CGP data with a PPI network} 
We applied SVD($OGL_0$, $L_0$) to the ``CGP+ PPI'' dataset consisting of the CGP gene expression and PPI network data. We set $k_u = 100$ to extract gene interactions, $k_v = 50$ to select 50 samples. Here we only focused on the first identified gene module, which contains 56 genes and 272 interactions. We first found that the subnetwork of identified module in the prior PPI network is dense (Fig. 3A). As we expected, the genes from this module contains a large number of linked genes in the prior PPI network (the degree sum of these genes is 6321). The identified 50 samples of this module are significantly related with some blood related cancers including AML (6 of 16), B cell leukemia (7 of 7), B cell lymphoma(7 of 10), Burkitt lymphoma(11 of 11), lymphoblastic leukemia(6 of 11), lymphoid neoplasm (5 of 11). On the other hand, these samples are specifically related with blood tissue (50 of 100). Moreover, this module is enriched in 77 different GO biological processes and 11 KEGG pathways, most of which are immune and blood related pathways including \emph{immune response-activating cell surface receptor signaling pathway}, \emph{immune response-regulating cell surface receptor signaling pathway} and \emph{immune response-activating signal transduction} (Fig. 3B).

Finally, we also applied $L_0$-SVD to extract a gene module with the same number of genes and samples for comparison. However, this module only contains 35 edges and most of the identified genes are isolated. We repeatedly sample 80 percent of genes and samples from the original matrix (i.e., the CGP data) for 10 times. For each sampled data matrices, we applied $OGL_0$-SVD and $L_0$-SVD to identify a gene module with the same settings above. We obtained an average of 382.8 (edges) for $OGL_0$-SVD, but an average of 35.6 (sum of edges) for $L_0$-SVD. These results indicate that $OGL_0$-SVD can identify gene modules with more connected edges in the prior PPI network by integrating edge-group structure.

\begin{table}[htbp]
\caption{The top gene modules identified by SVD($OGL_0$, $L_0$) and $L_0$-SVD from 12 different cancer-type gene expression datasets. \#edge indicates the number of edges of all module genes, and \#degree indicates the sum of degree of all module genes in the prior PPI network.}
\centering
\resizebox{\columnwidth}{!}{
\begin{threeparttable}
\begin{tabular}{p{2cm}|c|c|c|c|c}
  \hline
   Data& \#gene & \#edge\tnote{a} & \#edge\tnote{b}& \#degree\tnote{a} & \#degree\tnote{b} \\
  \hline
  BLCA-module &66& 22 & \textbf{135} & 1936 & 1800  \\
  \hline
  SKCM-module &65& 82 & \textbf{202} & 8348 & 2800  \\
  \hline
  LUAD-module &73& 25 & \textbf{173} & 1434 & 2948 \\
  \hline
  LUSC-module &67& 19 & \textbf{170} & 793  & 2813  \\
  \hline
  BRCA-module &79& 83 & \textbf{162} & 8493 & 5211 \\
  \hline
  HNSC-module &84& 23 & \textbf{159} & 1231 & 3598  \\
  \hline
  THCA-module &50& 52 & \textbf{382} & 6871 & 4699  \\
  \hline
  KIRC-module &66& 29 & \textbf{194} & 4848 & 10497 \\
  \hline
  LGG-module  &59&198 & \textbf{139} & 6048 & 7549  \\
  \hline
  PRAD-module &68& 69 & \textbf{178} & 7463 & 9210  \\
  \hline
  UCEC-module &55& 22 & \textbf{251} & 4512 & 14144 \\
  \hline
  CRC-module  &49&  1 & \textbf{384} & 1177 & 16620 \\
  \hline
\end{tabular}
\begin{tablenotes}
  \footnotesize
  \item[a] corresponds to $L_0$-SVD.
  \item[b] corresponds to SVD($OGL_0$, $L_0$).
\end{tablenotes}
\end{threeparttable}
}
\end{table}
\begin{figure*}[htbp]
  \centering
  \includegraphics[width=0.95\linewidth]{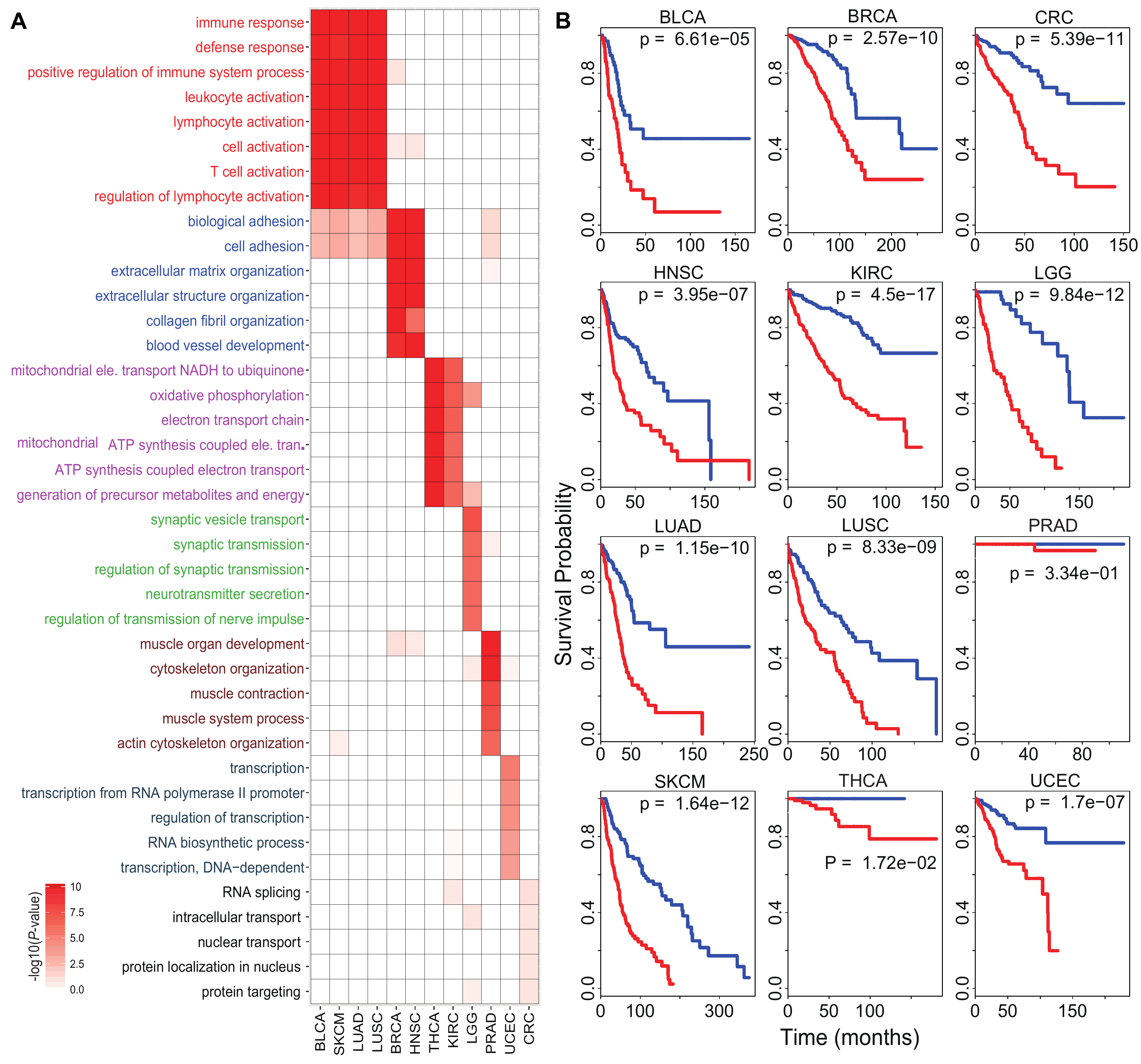}
  \caption{Results of ``TCGA + PPI'' dataset. (A) Top five enriched GOBP (Gene Ontology Biological Process) terms  of each gene module are shown where enriched scores were computed by -log10($p$-value) (where $p$-value represents the Benjamini-Hochberg adjusted $p$-value). To visualize the results, if -log10($p$-value) is greater than 10, it is set to 10. (B) Kaplan-Meier survival curves show overall survival of 12 different cancer types. $p$-values (or $p$s) were computed by log-rank test and `+' denote the censoring patients. }
\end{figure*}
\begin{table*}[htbp]
\centering
\caption{Summary of the top ten gene modules identified by SVD($OGL_0$, $OGL_0$) in the ``CGP+KEGG" dataset. \#Gene and \#Sample denote the number of genes and samples in each module; KEGG Term denotes the enriched KEGG pathways; Tumor and Tissue Type denote the enriched tumor and tissue types respectively.}
\begin{tabular}{l|l|l|p{6.5cm}|p{5cm}|p{2cm}}
  \hline
  ID& \#Gene & \#Sample & KEGG Term & Tumor Type & Tissue Type \\
  \hline
  1 & 221 & 100 & Ribosome; Primary immunodeficiency; B cell receptor signaling pathway; Intestinal immune network for iga production; Asthma & AML; B cell leukemia; B cell lymphoma; Burkitt lymphoma; lymphoblastic leukemia; Lymphoid neoplasm other & Blood \\
  \hline
  2 & 99 & 100 & DNA replication; Homologous recombination; Base excision repair; Mismatch repair; Nucleotide excision repair & B cell leukemia; Lung: Small cell carcinoma; Lymphoblastic leukemia; Lymphoblastic T cell leukaemia & Blood; Lung \\
  \hline
  3 & 200 & 100 & Ecm receptor interaction; Glycosaminoglycan degradation; Primary immunodeficiency; Glycosaminoglycan biosynthesis heparan sulfate; Arrhythmogenic right ventricular cardiomyopathy arvc & Glioma; lymphoblastic leukemia; Lymphoblastic T cell leukaemia; Osteosarcoma & Blood; Bone; CNS; Soft-tissue \\
  \hline
  4 & 216 & 100 & Pathogenic escherichia coli infection; Lysine degradation; Notch signaling pathway; Adherens junction; Rna degradation & AML; B cell leukemia; B cell lymphoma; Burkitt lymphoma; Lymphoblastic leukemia; Lymphoblastic T cell leukaemia; Lymphoid neoplasm other & Blood \\
  \hline
  5 & 83 & 100 & Allograft rejection; Type i diabetes mellitus; Asthma; Graft versus host disease; Intestinal immune network for iga production & AML; B cell leukemia; B cell lymphoma; Burkitt lymphoma; Hodgkin lymphoma; Lymphoblastic T cell leukaemia; Lymphoid neoplasm other; Myeloma & Blood \\
  \hline
  6 & 166 & 100 & Complement and coagulation cascades; Phenylalanine metabolism; Primary bile acid biosynthesis; Ppar signaling pathway; Steroid biosynthesis & Large intestine; Liver & GI tract \\
  \hline
  7 & 253 & 100 & Hematopoietic cell lineage; Acute myeloid leukemia; Fc epsilon ri signaling pathway; Fc gamma r mediated phagocytosis; Primary immunodeficiency & AML; B cell lymphoma; Lymphoblastic leukemia; lymphoblastic T cell leukaemia; lymphoid neoplasm other; oesophagus; upper aerodigestive tract & Blood; Upper aerodigestive \\
  \hline
  8 & 141 & 100 & Glutathione metabolism; Metabolism of xenobiotics by cytochrome p450; Steroid hormone biosynthesis; Ascorbate and aldarate metabolism; Arachidonic acid metabolism & Liver; lung: NSCLC: adenocarcinoma; Lung: NSCLC: squamous cell carcinoma; Oesophagus & GI tract; Lung; Upper aerodigestive \\
  \hline
  9 & 132 & 100 & Glycosaminoglycan biosynthesis chondroitin sulfate; Mismatch repair; N glycan biosynthesis; Glycosylphosphatidylinositol GPI anchor biosynthesis; Vibrio cholerae infection & Glioma; Lung: small cell carcinoma; Lymphoblastic leukemia; Neuroblastoma & CNS \\
  \hline
  10 & 193 & 100 & B cell receptor signaling pathway; Circadian rhythm mammal; Primary immunodeficiency; Fc gamma r mediated phagocytosis; Fc epsilon ri signaling pathway & AML; B cell lymphoma; Glioma; lung: small cell carcinoma; Lymphoblastic leukemia; Lymphoid neoplasm other & Blood \\
  \hline
\end{tabular}
\end{table*}

\subsection{Application to TCGA data with a PPI network}
We next applied SVD($OGL_0$, $L_0$) to analyze the gene expression data of 12 different cancers including BLCA, BLCA, LUAD, LUSC, BRCA, HNSC, THCA, KIRC, LGG, UCEC, and CRC. Note that SVD($OGL_0$, $L_0$) was tested independently for each cancer to identify a important gene module across a set of tumor patients. SVD($OGL_0$, $L_0$) uses overlapping group $L_0$-norm penalty for gene interaction selection with $k_u = 100$ (i.e. identify 100 edge-groups) and $L_0$-norm penalty for sample-variable selection with $k_v = 50$ (i.e., 50 samples). Thus, we obtained 12 gene modules for 12 cancer types with about 65 genes on average (TABLE 3). For convenience, we defined the identified module of `X' cancer-type as X-module (TABLE 3). For comparison, we also enforced $L_0$-SVD to identify a gene module with the same number of samples and genes for each cancer type. As we expected, all modules identified by SVD($OGL_0$, $L_0$) contain more edges than the ones identified by $L_0$-SVD.

Interestingly, we also find that most of the module genes identified by SVD($OGL_0$, $L_0$) are with higher degree in the prior PPI network than those by $L_0$-SVD without using prior information (TABLE 3), indicating that these genes with higher degree tend to be related with cancer. In addition, we observed a overlapping pattern, containing four cancer-specific modules (BLCA, SKCM, LUAD and LUSC cancers which are related with skin/epidermis tissue) sharing many common genes. Subsequent functional analysis shows these common genes are enriched in some important immune-related pathways (Fig. 4A). Taken together of 12 cancer-type gene modules, we find 412 enriched GO biological processes (GO BPs, or GOBPs) and 48 KEGG pathways with Benjamin corrected $p$-value $<$ 0.05. Interestingly, some important caner-related KEGG pathways are discovered in both BRCA-module and HNSC-module including \emph{ECM-receptor interaction} (with $p$-value = 4.3e-18 in BRCA-module and $p$-value = 2.8e-22 in HNSC-module), \emph{Focal adhesion} (with $p$-value = 3.2e-16 in BRCA-module and $p$-value = 3.5e-22 in HNSC-module), and \emph{Pathways in cancer} (with $p$-value = 5.0e-03 in BRCA-module and $p$-value = 2.3e-04 in HNSC-module) (Fig. 4A). Some important immune related pathways in both BLCA, SKCM, LUAD and LUSC-module including \emph{Cell adhesion molecules}, \emph{Hematopoietic cell lineage}, \emph{Natural killer cell mediated cytotoxicity}, \emph{Primary immunodeficiency}, and \emph{T cell receptor signaling pathway}. This is consistent with previous studies that lung cancer is related to immune response \cite{roepman2009immune}. In addition, several KEGG pathways are found in THCA-module including \emph{Oxidative phosphorylation} ($p$ = 1.1e-59), \emph{Parkinson's disease} ($p$ =7.9e-52), \emph{Alzheimer's disease} ($p$ =5.8e-48) and \emph{Huntington's disease} ($p$ = 1.8e-46). For clarity, the top five enriched GO BPs of each cancer module are shown (Fig. 4A). Several GO BPs are common in many cancers, while most of them are specific to certain cancers (Fig. 4A).

Moreover, we find that those patients of 11 cancers (BLCA, BLCA, LUAD, LUSC, BRCA, HNSC, THCA, KIRC, LGG, UCEC, CRC) can be effectively separated into two risk-groups, which are significantly different with survival outcome ($p$-value $<$ 0.05, log-rank test) (Fig. 4B). All the results showed that our method by integrating interaction structure could identify more biologically relevant gene modules, and improve their biological interpretations.

\subsection{Application to CGP data with KEGG Pathways}
We applied SVD($OGL_0$, $L_0$) to the ``CGP + KEGG''. We set $k_u = 5$ (i.e., five pathways), $k_v = 100$ (i.e. 100 samples) in SVD($OGL_0$, $L_0$) for convenience. Since the identified top ten pair of singular vectors explained more than 60\% of the variance, we focused on the top ten pair singular vectors to extract ten gene functional modules (TABLE 4).

For each gene module, we computed the overlap significance between its sample set and each tissue or cancer class using a hypergeometric test, implemented via $R$ function $phyper$. Those cancer types with a few samples ($\leq 5$) are ignored. We find that each module is significantly related to at least one cancer or tissue type, indicating that they indeed be biologically relevant.

It is worth noting that some cancer/tissue subtype-specific KEGG pathways are discovered. We summarized all key messages of the identified modules in TABLE 4. For example, we find that all the samples of module 1 belong to blood tissue, and most of samples of module 1 are significantly enriched in some lymphoid-related cancers. Interestingly, the corresponding five blood-specific KEGG pathways (including \emph{ribosome}, \emph{primary immunodeficiency}, \emph{b cell receptor signaling pathway}, \emph{intestinal immune network for iga production}, and \emph{asthma}) are related to lymphoma. Another example module 6 is specifically related to large intestine, liver cancer and GI tract tissue with important KEGG pathways including \emph{complement and coagulation cascades}, \emph{phenylalanine metabolism}, \emph{primary bile acid biosynthesis}, \emph{ppar signaling pathway}, \emph{steroid biosynthesis}. These results provide a new way to understand and study the mechanisms between different tissues and cancers.

\section{Discussion and conclusion}
Inferring blocking patterns from high-dimensional biological data is still a central challenge in computational biology. On one hand, we aim to identify some gene subsets, which are co-expressed across some samples in the corresponding gene expression data. On the other hand, we expect that these identified gene sets belong to some biologically meaningful groups such as functional pathways. To this end, we propose group-sparse SVD models to identify gene blocking patterns (modules) via integrating gene expression data and prior gene knowledge.

We note that the concept of group sparse has also been introduced into some related models such as the non-negative matrix factorization (NMF) model \cite{Kim2012Group}. To our knowledge, non-overlapping group Lasso was applied as the penalty onto NMF. Obviously, our models are very different from the group-sparse NMF model. First, we used a more direct non-overlapping sparse penalty with $L_0$-norm penalty into the SVD model and propose a group sparse SVD with group $L_0$-norm penalty ($GL_0$-SVD). More importantly, we prove the convergence of the $GL_0$-SVD algorithm. Second, since the non-overlapping group structure in group Lasso or group $L_0$-norm penalty limits their applicability in practice. Several works have studied the (overlapping) group Lasso in regression tasks. However, little work focus on developing structured sparse matrix factorization models with overlapping group structure. We propose two overlapping group SVD models ($OGL_1$-SVD and $OGL_0$-SVD), and discuss their convergence issue. Computational results of high-dimensional gene expression data show that our methods could identify more biologically relevant gene modules and improve their biological interpretations than the state-of-the-art sparse SVD methods. Moreover, we expect that our methods could be applied to more high-dimensional biological data such as single cell RNA-seq data, and high-dimensional data in other fields. In the future, it will be valuable to extend current concept into structured sparse tensor factorization model for multi-way analysis of multi-source data.

\section*{Acknowledgement}
Shihua Zhang and Juan Liu are the corresponding authors of this paper. Wenwen Min would like to thank the support of the Academy of Mathematics and Systems Science, CAS during his visit. This work has been supported by the National Natural Science Foundation of China [11661141019, 61621003, 61422309, 61379092]; Strategic Priority Research Pro-gram of the Chinese Academy of Sciences (CAS) [XDB13040600]; National Ten Thousand Talent Pro-gram for Young Top-notch Talents; Key Research Pro-gram of the Chinese Academy of Sciences [KFZD-SW-219]; National Key Research and Development Program of China [2017YFC0908405]; CAS Frontier Science Re-search Key Project for Top Young Scientist [QYZDB-SSW-SYS008].

\bibliographystyle{IEEEtran}
\balance
\bibliography{References_gSVD}
\end{document}